%% file: main.tex
\documentclass[11pt]{article}

\usepackage[final]{acl}

\usepackage{times}
\usepackage{latexsym}

\usepackage[T1]{fontenc}

\usepackage[utf8]{inputenc}

\usepackage{microtype}

\usepackage{inconsolata}

\usepackage{graphicx}
\graphicspath{{figures/}}

\usepackage{caption}
\usepackage{subcaption}

\usepackage{amsmath}
\usepackage{amssymb}

\usepackage{booktabs}
\usepackage{multirow}
\usepackage{tabularx}
\usepackage[table]{xcolor}

\usepackage{xspace}
\usepackage{cuted}
\usepackage{titletoc}

\newcommand{\pname}{{ORLF}\xspace}
\newcommand{\dname}{{TKFQA}\xspace}

\title{
Counterfactual Benchmarking and Training for Factuality Consistency
and Order-Robust Grounded Reasoning in LLMs over Heterogeneous Knowledge
}

\author{
\textnormal{Shibo Chu}\textsuperscript{1},
\textnormal{Yuze Liu}\textsuperscript{2},
\textnormal{Tiehua Zhang}\textsuperscript{1}\ensuremath{^{\dagger}},
\textnormal{Zhishu Shen}\textsuperscript{3},
\\
\textnormal{Lianghua He}\textsuperscript{1},
\textnormal{Haofen Wang}\textsuperscript{1},
\textnormal{Zhijun Ding}\textsuperscript{1}
\\[3pt]
\textsuperscript{1}Tongji University, Shanghai, China
\\
\textsuperscript{2}Swinburne University of Technology, Melbourne, Australia
\\
\textsuperscript{3}Wuhan University of Technology, Wuhan, China
}

\begin{document}

\maketitle

\begingroup
\renewcommand{\thefootnote}{\fnsymbol{footnote}}
\footnotetext[2]{Corresponding author.}
\endgroup

\input{sections/00_Abstract}

\input{sections/01_Introduction}

\input{sections/02_Related_work}

\input{sections/03_Dataset}

\input{sections/04_Method}

\input{sections/05_Experiments}

\input{sections/Limitations}

\bibliography{custom}

\appendix

\input{sections/Appendix}
\end{document}

%% file: sections/00_Abstract.tex
\begin{abstract}
Large language models (LLMs) have increasingly supported response generation grounded in user-provided knowledge spanning heterogeneous structures. However, existing benchmarks provide limited assessment of whether LLMs can faithfully perform multi-hop reasoning chains across such knowledge contexts while remaining robust to variations in their input order. We introduce \dname, a factuality consistency benchmark comprising 10,130 question-answering (QA) pairs grounded in tables, texts, and knowledge graphs (KGs). Each example is constructed from an explicit counterfactual reasoning chain, enabling the joint evaluation of answer correctness, reasoning-chain accuracy, and robustness to different input-order. An extensive evaluation of 14 open- and closed-source LLMs reveals that state-of-the-art models exhibit limited reasoning-chain accuracy and remain sensitive to variations in the input order of heterogeneous knowledge contexts. To address these limitations, we propose \pname, an LLM-agnostic training framework that models cross-context topological relations through knowledge-specific latent vectors. \pname integrates context-wise position encoding, a latent-bridge attention mask, and topological knowledge bias to preserve knowledge-specific bias and encode topological semantics. Experiments across four LLM backbones show that \pname outperforms competitive training-free and LoRA-based baselines, improving average Exact Match and Reasoning-Chain Accuracy by 2.15\% and 4.29\%, respectively, while reducing order-induced performance standard deviation by 0.04\% to 3.01\%. 
\end{abstract}

%% file: sections/01_Introduction.tex
\section{Introduction}
\label{sec:introduction}
Large language models (LLMs) have achieved substantial progress across diverse domains, including medical diagnosis~\cite{zhang2026medtvt}, financial risk control~\cite{chen2026tasks}, and software development~\cite{liang2022astbert}. A widely adopted usage paradigm involves conditioning LLMs on external knowledge contexts to generate responses grounded in the provided information~\cite{zhao2026retrieval}. Nevertheless, as LLMs are deployed in increasingly complex task settings, effective grounded reasoning over user-provided knowledge contexts remains a fundamental challenge~\cite{jacovi2025facts,liu2025comprehensive,liu2026structure}.  A surge of datasets have been proposed to evaluate the grounded reasoning abilities of LLMs from different perspectives, including multi-hop  reasoning~\cite{yang2018hotpotqa,wu2025cofca} and reasoning over diverse structured knowledge contexts.~\cite{chen2020hybridqa,lei2023s3hqa}. However, existing benchmarks largely overlook factuality consistency~\cite{jacovi2025facts} during LLM inference, which refers to whether generated responses are faithfully grounded in user-provided knowledge contexts.

Addressing this gap, we propose a factuality evaluation benchmark for multi-hop grounded reasoning, namely \dname, which consists of 10,130 question-answering (QA) pairs associated with heterogeneous knowledge contexts, including tables, texts, and knowledge graphs (KGs). Across knowledge contexts with diverse structures, the carefully constructed counterfactual chain provides the reference multi-hop reasoning sequence for deriving the correct final answer. The counterfactual entities within this chain enable assessment of whether LLMs faithfully ground their reasoning in the input-provided knowledge, rather than relying solely on their internal parametric knowledge~\cite{afzal2025knowing}. Solving problems in \dname requires an LLM to perform grounded reasoning over heterogeneous knowledge contexts, capture relations among different knowledge-context pairs, and conduct multi-hop reasoning to derive the final answer. In addition, during inference, variations in the input order of heterogeneous knowledge contexts can lead to fluctuations in model performance~\cite{guan2025order}.

To address these problems, we further propose \pname, an LLM-agnostic training framework that capture topological relations among heterogeneous knowledge contexts through knowledge-specific latent vectors, thereby maintaining factual consistency and mitigating sensitivity to input-order variations during multi-hop grounded reasoning. The framework consists of three key components: context-wise position encoding (CPE), which adapts positional encoding within each knowledge context; latent-bridge attention mask (LBAM), which enables latent vectors to attend across different knowledge contexts; and topological knowledge bias (TKB), which adaptively encodes topological semantics.

In summary, the main contributions are summarized as follows:
\begin{itemize}
\item We introduce \dname, a factuality consistency evaluation benchmark for multi-hop grounded reasoning, comprising 10,130 QA pairs grounded in tables, text, and KG. Each QA pair is constructed from an explicit counterfactual reasoning chain, enabling the evaluation of answer correctness, reasoning-chain accuracy, and robustness to knowledge-context input-order variations.
\item We propose \pname, an LLM-agnostic training framework that captures topological relations among heterogeneous knowledge contexts through knowledge-specific latent vectors. It integrates context-wise position encoding (CPE), a latent-bridge attention mask (LBAM), and topological knowledge bias (TKB) to preserve factuality consistency and reduce sensitivity to input-order variations.
\item We conduct comprehensive inference experiments on \dname involving 14 state-of-the-art open- and closed-source LLMs. The results reveal that these models remain limited in their ability to understand heterogeneous knowledge contexts and perform robust multi-hop grounded reasoning. Further experiments demonstrate that the proposed \pname consistently improves both answer accuracy and reasoning-chain accuracy while substantially enhancing robustness to input-order variations across different LLM backbones.
\item We have publicly released the raw benchmark data and source code at https://github.com/xq7m4k9/a8v3n2 to facilitate reproducibility and support further research in this field.

\end{itemize}

%% file: sections/02_Related_work.tex
\section{Related Work}
\label{sec:related-work}
\paragraph{Factuality Evaluation in Grounded Reasoning.}
Existing benchmarks for evaluating factuality primarily focus on grounded reasoning within a single knowledge structure and do not explicitly assess the correctness of reasoning chains generated during LLM inference. RAGTruth~\cite{niu2024ragtruth} and FACTS~\cite{jacovi2025facts} assess whether generated responses are supported by retrieved or provided documents, while HaluEval~\cite{li2023halueval} and FActScore~\cite{min2023factscore} focus on hallucination detection and factual support for generated claims. However, these benchmarks largely evaluate output-level factuality in homogeneous knowledge contexts, while the correctness of multi-hop reasoning chains across heterogeneous knowledge structures insufficiently explored.  In contrast, \dname evaluates grounded reasoning over tables, texts, and knowledges graphs using counterfactual chains, enabling joint assessment of answer correctness, reasoning-chain accuracy, and input-order robustness.

\paragraph{Grounding LLMs in Structured Knowledge.}
Existing methods to grounded reasoning can be broadly categorized into training-free and LoRA-based methods. Training-free approaches, such as CoT-D~\cite{wang2024chain} and ReAct~\cite{yao2022react}, elicit intermediate reasoning without updating model parameters. LoRA-based methods introduce specialized attention mechanisms to improve cross-evidence interaction and multi-hop reasoning. For example, TXH employs extra-hop attention to construct globally contextualized representations, while PMFT~\cite{huang2025masking} investigates bidirectional attention under context permutations. Nevertheless, these methods primarily focus on homogeneous knowledge structures or generic cross-context interactions and do not explicitly model topological dependencies among heterogeneous knowledge contexts to preserve factuality consistency. These limitations are addressed by our proposed framework.

%% file: sections/03_Dataset.tex
\section{\dname}
\label{sec:dataset}

\begin{figure*}[t]
    \centering
    \vspace{-3mm}
    \includegraphics[
        width=\textwidth,
        keepaspectratio
    ]{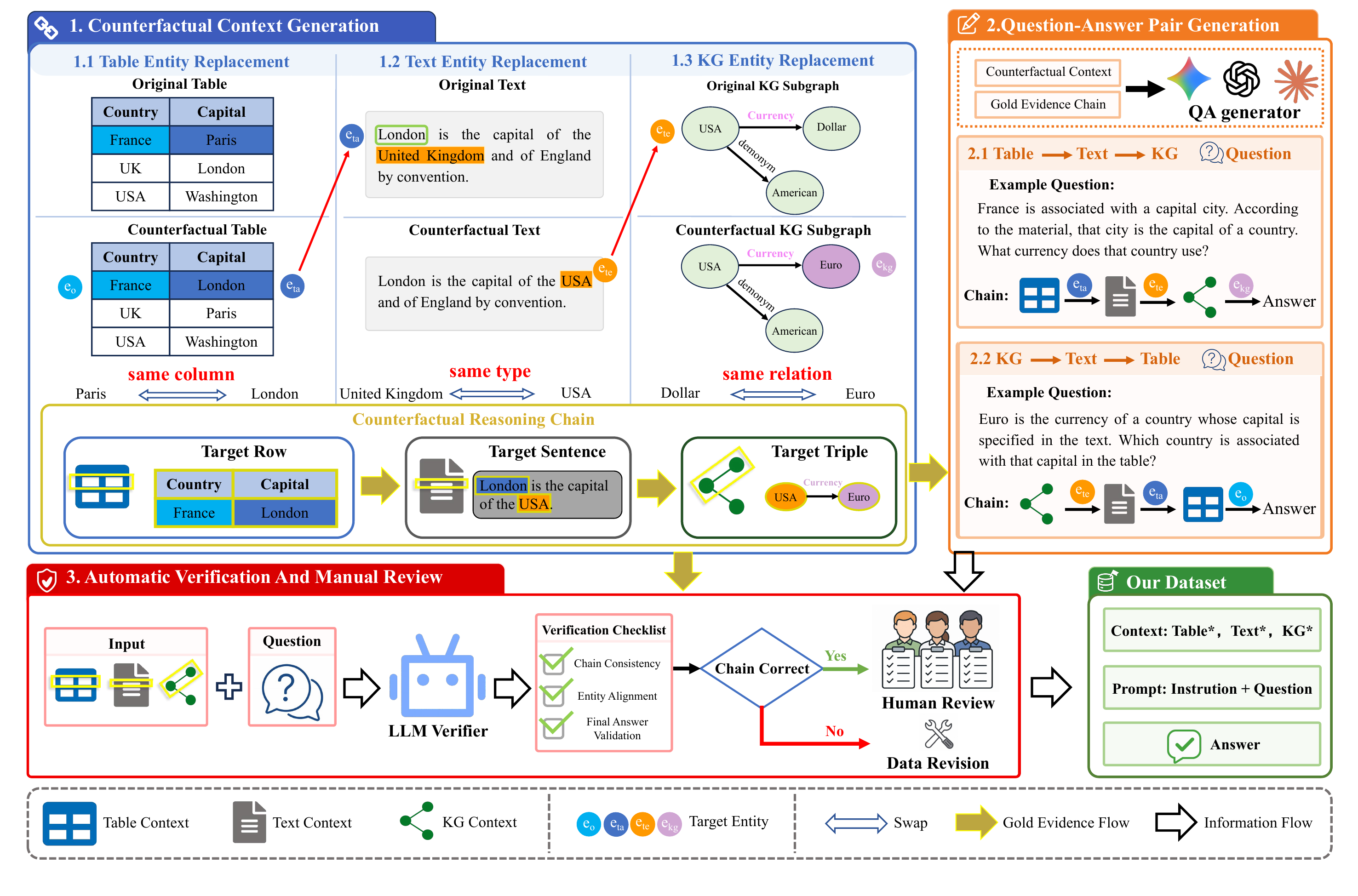}
    \vspace{-10mm}
    \caption{
            Overview of the {\dname} construction pipeline. We first construct counterfactual contexts and a counterfactual evidence chain spanning a table, a textual passage, and a KG subgraph. We then generate questions, followed by automatic verification and manual review.
            }
    \label{fig:dataset_construction}
    \vspace{-5mm}
\end{figure*}


\label{sec:dataset_construction}
We first collect source data from Wikipedia and the Wikidata KG~\citep{vrandecic2014wikidata}, selecting Wikipedia tables with non-empty column headers and at least one entity-linked cell. Then, we obtain textual passages linked to the selected tables and retrieve relevant KGs from Wikidata for the entities mentioned in these passages~\citep{chen2020hybridqa}. These texts, tables, and KGs are used as source materials for subsequent counterfactual context construction.


\paragraph{Counterfactual Context Generation.}
\label{sec:counterfactual_context_generation}
Although LLMs have demonstrated strong reasoning capabilities across a wide range of tasks~\cite{liu2025grl,jiangdivide}, prior studies~\cite{tang2024minicheck,xu2024pride} suggest that they may rely on internal parametric knowledge rather than external knowledge provided in user queries when generating grounded responses. To systematically investigate this issue, we design a counterfactual context generation process consisting of three steps: table entity replacement, text entity replacement, and KG entity replacement, as illustrated in Part 1 of Figure~\ref{fig:dataset_construction}. 



\noindent\textbf{1) Table Entity Replacement.}
We select a row from a original Wikipedia table as the target row and replace one of its entity-linked cells with another entity of the same semantic category from the corresponding column. Here, the semantic category refers to the entity type, such as country or city. Concretely, as illustrated in Part 1.1 of Figure~\ref{fig:dataset_construction}, the row containing the entity “France” is selected, and “London” in the \textit{Capital} column is replaced with “Paris”. The replacement entity inserted into the target row is denoted as $e_{ta}$, while another entity from the same row is denoted as $e_o$. The counterfactual table obtained after entity replacement is denoted as $\Tilde{T}$.

\paragraph{2) Text Entity Replacement.}
Based on the replacement entity $e_{ta}$ in the counterfactual table, text entity replacement is performed on the corresponding Wikipedia passage, as shown in Part 1.2 of Figure~\ref{fig:dataset_construction}. Specifically, another entity in the text is replaced with $e_{te}$, which belongs to the same semantic category. For example, in the Wikipedia passage associated with $e_{ta}$, “London”, the entity “United Kingdom” is replaced with the entity $e_{te}$, “USA”. We denote the counterfactual text obtained after entity replacement as $\Tilde{P}$.

\noindent\textbf{3) KG Entity Replacement.}
Similarly, we retrieve the entity $e_{te}$ from the counterfactual text in the Wikipedia KG and construct a KG subset based on its one-hop neighbors. Within this subset, one neighboring entity is randomly selected and replaced with a replacement entity $e_{kg}$ from the same semantic category. The resulting counterfactual KG after entity replacement is denoted by $\Tilde{G}$. We generate a counterfactual reasoning chain across heterogeneous structured knowledge, including tables, texts, and KGs, linked by the replacement entities $e_{ta}$, $e_{te}$, and $e_{kg}$. Mathematically, the counterfactual reasoning chain is denoted as $\tilde{\mathcal{C}} = \langle\tilde{T}, \tilde{P}, \tilde{G}\rangle(e_o, e_{ta}, e_{te}, e_{kg})$, with an example provided in \textbf{Appendix~\ref{app:dataset_example}}. Details of the two subsequent stages—Question-Answer Pair Generation and Automatic Verification and Manual Review—are provided in \textbf{Appendix~\ref{app:additional_construction}} and illustrated in Parts 2 and 3 of Figure~\ref{fig:dataset_construction}, respectively.

\paragraph{Benchmark Statistics.}
\label{sec:dataset_statistics}
\input{tables/dataset_statistics}

Table~\ref{tab:dataset_statistics} summarizes the statistics of \dname, which comprises 10,130 QA pairs, each paired with a corresponding heterogeneous structured knowledge context across tables, texts, and KGs, generated from 5,065 counterfactual reasoning chains.We randomly split the corpus into training, development, and test sets at a ratio of 8:1:1, with the token lengths of the questions, answers, and heterogeneous structured knowledge contexts reported in Table~\ref{tab:dataset_statistics}.


\noindent\textbf{Evaluation Metrics.}

\noindent\textbf{1) Response Correctness Evaluation.} We adopt Exact Match (EM) to measure the consistency between the LLM response and the reference answer.

\noindent\textbf{2) Reasoning-Chain Correctness Evaluation.} We provide a prompt in \textbf{Appendix~\ref{app:rca}} that explicitly instructs LLMs to generate reasoning chains on \dname. The generated chains are then compared with the provided counterfactual reasoning chains, and Reasoning-Chain Accuracy (RCA) is used to measure their similarity, with a focus on entity-level consistency between the two chains.

\noindent\textbf{3) Input-Order-Aware Robustness Correctness Evaluation.}
Previous studies~\cite{huang2025masking} indicate that variations in the input order of heterogeneous knowledge contexts within prompts can lead to performance fluctuations in multi-hop grounded-reasoning QA tasks. We adopt Order Standard Deviation (O. Std.) to quantify performance variation under different input orders of knowledge contexts within prompts. Mathematically, we define $\Omega(T, P, G) = \{T\text{-}P\text{-}G, T\text{-}G\text{-}P, P\text{-}T\text{-}G, P\text{-}G\text{-}T, G\text{-}P\text{-}T, G\text{-}T\text{-}P\}$ as the set of all possible input orders of the heterogeneous knowledge contexts. For example, $T\text{-}P\text{-}G$ denotes that the table context, text context, and KG context are sequentially placed in the prompt. \dname requires LLMs to ground their predictions in heterogeneous knowledge contexts, model the relations among different context types, and perform multi-hop reasoning to derive the final answer. Moreover, model performance may vary substantially with the ordering of these knowledge contexts during inference.



%% file: tables/dataset_statistics.tex
\begin{table}[t]
    \centering
    \small
    \setlength{\tabcolsep}{3.8pt}
    \renewcommand{\arraystretch}{1.08}

    \begin{tabularx}{\columnwidth}{@{}Xrrrr@{}}
        \toprule
        \textbf{Split}
        & \textbf{Train}
        & \textbf{Val}
        & \textbf{Test}
        & \textbf{Total} \\
        \midrule

        \# Samples
        & 8,106
        & 1,012
        & 1,012
        & 10,130 \\

        \midrule
        \multicolumn{5}{c}{\textbf{\# Avg. Tokens}} \\
        \midrule

        Table
        & 856.45
        & 782.06
        & 800.72
        & 844.33 \\

        Text
        & 3,665.10
        & 3,327.82
        & 3,627.96
        & 3,638.26 \\

        KG
        & 3,704.24
        & 3,973.04
        & 3,794.92
        & 3,733.60 \\

        Question
        & 68.80
        & 68.87
        & 68.33
        & 68.74 \\

        Answer
        & 6.06
        & 6.18
        & 5.66
        & 6.01 \\

        All
        & 8,322.85
        & 8,180.15
        & 8,319.80
        & 8,313.15 \\

        \bottomrule
    \end{tabularx}
    \vspace{-3mm}

    \caption{
        Statistics of TKFQA across the train, validation, and test splits.
    }
    \vspace{-5mm}
    \label{tab:dataset_statistics}
\end{table}

%% file: sections/04_Method.tex
\section{Methodology}
\label{sec:method}
\begin{figure*}[t]
\vspace{-5mm}
    \centering
    \includegraphics[
        width=\textwidth,
        keepaspectratio
    ]{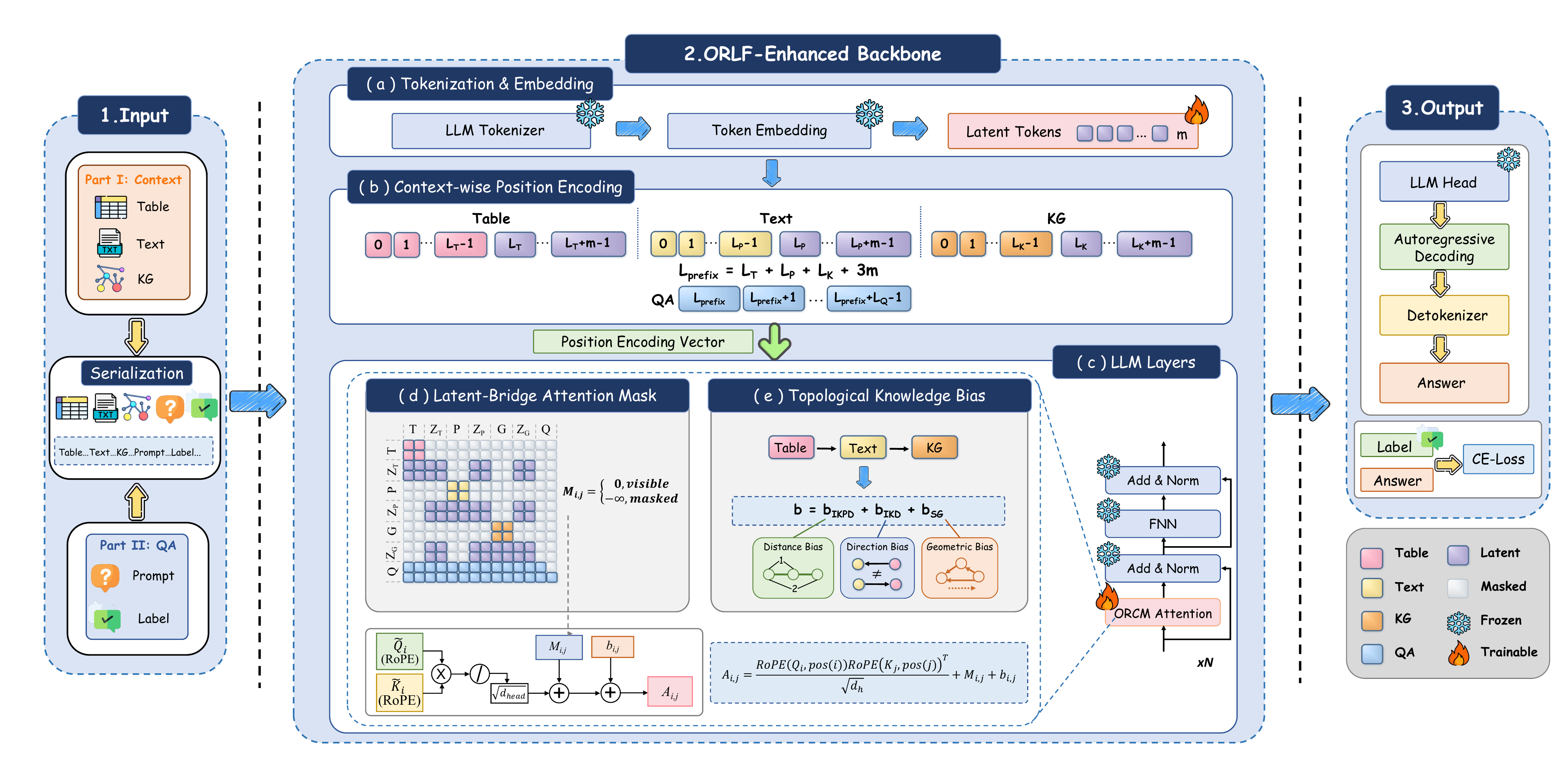}
    \vspace{-10mm}
    \caption{
        Overview of the proposed \pname framework.} 
        
    
    \label{fig:orcm_overview}
    \vspace{-5mm}
\end{figure*}

To address the challenges in \dname, we first formulate the multi-hop grounded reasoning problem for \dname and then introduce a novel fine-tuning framework for addressing this problem through \textbf{O}rder-\textbf{R}obust \textbf{L}atent \textbf{F}usion, namely \pname, as shown in Figure~\ref{fig:orcm_overview}.

\subsection{Problem Formulation}
\label{sec:problem_formulation}
Let $D_{tr} = {((q, a), C_{\omega})}$ denote the training dataset of \dname, where $(q, a)$ represents a QA pair and $C_{\omega} = \{T, P, G\}_{\omega}$ denotes the corresponding heterogeneous knowledge context. Specifically, $\omega \in \Omega(T, P, G)$ denotes the input order of the heterogeneous knowledge contexts within the prompt. The embedded sequence input to the LLM $\emph{M}$ can be denoted as follows:
\begin{equation}
    \emph{X} = E((q,a),C_{\omega}) = [\emph{X}_{\omega}^{C}, \emph{X}^{q}, \emph{X}^{a}]
\end{equation}where $E(\cdot)$ denotes the tokenization and embedding operation, and $\emph{X} = [\emph{X}_{\omega}^{C}, \emph{X}^{q}, \emph{X}^{a}]$ denotes the input vector corresponding to one training sample. Specifically, $\emph{X}_{\omega}^{C}$, $\emph{X}^{q}$, and $\emph{X}^{a}$ denote the embedded representations of the knowledge context $C_{\omega}$, question $q$, and answer $a$, respectively. We aim to train the LLM $\emph{M}$ on the training dataset $D_{tr}$ by minimizing the cumulative log-likelihood loss of the LLM generated response, which can be formulated as follows:
\begin{equation}
    \mathcal{L}_{\pi}
    =
    -
    \sum_{i=1}^{|D_{tr}|}\sum_{j=1}^{|y^i|}
    \log
    P_{\theta}
    \left(
        y_j^i
        \mid
        \emph{X},
        y_{<j}^i
    \right)
    \label{eq:training_objective}
\end{equation}where $|D_{tr}|$ denotes the number of training samples, and $|y^{i}|$ denotes the number of tokens in the LLM-generated answer for the $i$-th sample.

\subsection{\pname}
We first initialize a set of trainable latent vectors, $\mathcal{Z} = \{\emph{z}^c\}^{c \in C}$, and append each vector to the corresponding input embedding vectors $\emph{X}_{\omega}^{C} = \{\emph{x}_{\omega}^{c}\}^{c \in C}$ for each structured knowledge type. Mathematically, this process is represented as follows:
\begin{equation}
    \begin{aligned}
        \emph{z}^{c}
        &=
        [z^{c}_{0},\ldots,z^{c}_{m-1}], \\
        \Tilde{\emph{x}}^{c}_{\omega}
        &=
        [\emph{x}^{c}_{\omega};\emph{z}^{c}]
    \end{aligned}
    \qquad
    c\in C
    \label{eq:latent_tokens}
\end{equation} For each knowledge structure type $c \in C$, the trainable latent vector $\emph{z}^{c}\in\mathcal{R}^m$ captures structure-specific attention shifts during LLM generation in multi-hop reasoning, enabling better integration of heterogeneous user-provided knowledge.


\noindent\textbf{Context-wise Position Encoding.} 
To mitigate the model’s sensitivity to the input order of heterogeneous knowledge contexts~\citep{su2021roformer}, we introduce CPE, in which positional indices are reset and assigned independently within each knowledge context type, as shown in Part (b) of Figure~\ref{fig:orcm_overview}. The position encoding vector for the embedded representation of the knowledge context $\Tilde{\emph{X}^C} = \{\Tilde{\emph{x}}^c_{\omega}\}^{c\in C}$ are assigned as follow:
\begin{equation}
    \begin{aligned}
        \operatorname{pos}(\emph{x}^{c}_{\omega})
        &=
        [0,\ldots,L_c-1], \\
        \operatorname{pos}(\emph{z}^{c})
        &=
        [L_c,\ldots,L_c+m-1]
    \end{aligned}
    \label{eq:source_positions}
\end{equation}where \(\operatorname{pos}(\cdot)\) denotes the function that assigns positional encoding vectors to the embedded input representations, and $L_c$ represents the length of the embedded representation $\emph{x}^c_\omega$. The total length of the embedded representation of the knowledge context $\Tilde{\emph{X}_\omega^C}$ is defined as follows:
\begin{equation}
    L_{\mathrm{prefix}}
    =
    \sum_{c\in C}(L_c+m)
    \label{eq:prefix_length}
\end{equation} The length of the remaining components in the prompt template, including the question, answer, and instruction tokens placed after the knowledge context, is denoted as $L_Q$. Details of the prompt used during training are provided in \textbf{Appendix~\ref{app:standard_prompt}}.

\noindent\textbf{Latent-Bridge Attention Mask.} Although CPE mitigates input-order-induced shifts in positional encoding vectors, causal attention still creates asymmetric visibility across different structured knowledge-context segments~\citep{haviv2022transformer,egressy2025setllm}. To this end, we introduce the Latent-Bridge Attention Mask, which allows trainable latent representations to attend to tokens across different knowledge-context segments during decoding, thereby enabling controlled cross-structure information exchange in attention-score computation. In contrast, the original knowledge-context tokens are restricted to attending only to tokens within their corresponding knowledge type, as shown in Part (e) of Figure~\ref{fig:orcm_overview}. Mathematically, the attention mask is formulated as follows:
\begin{equation}
M_{ij}=
\begin{cases}
0, & i,j\in\mathcal{X}^c,\\
0, & i\in\mathcal{Z}^c,\;
     j\in\mathcal{X}^c\cup\mathcal{Z},\\
0, & i\in\mathcal{Q}\cup\mathcal{Y},\;j\le i,\\
-\infty, & \text{otherwise}.
\end{cases}
\label{eq:latent_mediated_mask}
\end{equation} Here, $\mathcal{X}^c$ and $\mathcal{Z}^c$ denote the index sets in the attention mask associated with the context-token embeddings $\emph{x}^c_\omega$ and the trainable latent representation $\emph{z}^c$, respectively. We define the index set of all latent vector $\emph{z}^c$ as $\mathcal{Z}=\bigcup_{c \in \mathcal{C}}\mathcal{Z}^c$, and use $\mathcal{Q}$ and $\mathcal{Y}$ to denote the index sets of the embedded question and answer representations, respectively.


\noindent\textbf{Topological Knowledge Bias.} Previous studies~\cite{vajda2026teaching,al2026dagger} indicate that knowledge contexts in multi-hop grounded-reasoning tasks may exhibit underlying relational dependencies. In \dname, for example, heterogeneous knowledge contexts are constructed from counterfactual reasoning chains, thereby naturally forming cross-structure topological relations, such as $\mathrm{Table} \rightarrow \mathrm{Text} \rightarrow \mathrm{KG}$. To further enable the latent vectors to capture topological relations across heterogeneous knowledge-context structures, we introduce TKB into the attention computation process. Specifically, this bias is constructed from three perspectives to capture topological semantics: inter-knowledge path distance, inter-knowledge direction, and spectral geometry. During attention computation in the $\ell$-th Transformer layer and the $h$-th attention head of the LLM, the TKB between the latent vector of knowledge type $u \in C$, which serves as the query vector, and the latent vector of knowledge type $v \in C$, which serves as the key vector, is formulated as follows:
\begin{equation}
b^{(\ell,h)}_{u, v}
=
b_{\mathrm{IKPD}}^{(\ell,h)}(u,v)
+
b_{\mathrm{IKD}}^{(\ell,h)}(u,v)
+
b_{\mathrm{SG}}^{(\ell,h)}(u,v).
\label{eq:topological_bias}
\end{equation} Specifically, for $b_{\mathrm{IKPD}}^{(\ell,h)}(u,v)$, we use a learned lookup table to encode the shortest-path distance $D_{hops}$, which measures the number of hops between different knowledge contexts, formulated as follows:
\begin{equation}
    b_{\mathrm{IKPD}}^{(\ell,h)}(u, v) = f_{IKPD}^{(\ell, h)}(D_{hops}(u, v);\theta)
\end{equation}where $\theta\in\mathcal{R}^{|C|\times|C|\times|C|}$ is learnable parameter, with $b_{\mathrm{IKPD}}^{(\ell,h)}(u, u) = 0$ set to 0. Similarly, the directional relation between each pair of knowledge contexts is also learned by $f_{IKD}^{(\ell, h)}(\mathcal{I}(u, v);\theta)$, where $\mathcal{I}(u,v)$ is an indicator function that specifies whether knowledge context $u$ is directed to knowledge context $v$, and $\theta \in \mathcal{R}^{|C| \times |C| \times |C|}$ is a learnable parameter. Meanwhile, we use a Magnetic Laplacian kernel~\citep{zhang2021magnet} to encode spectral geometric relations in the knowledge-context topology into $b_{\mathrm{SG}}^{(\ell,h)}(u,v)$ through a permutation-invariant transformation of the kernel eigenvalues. The three bias terms are incorporated into the raw attention scores, as illustrated in Figure~\ref{fig:orcm_overview}(e). The final attention scores are computed as follows:
\begin{align}
A_{ij}^{(\ell,h)}
\!\!=\!\!{}&
\frac{
    \operatorname{RoPE}(\emph{Q}_i^{(\ell,h)}\!\!,pos(i))
    \operatorname{RoPE}(\emph{K}_j^{(\ell,h)}\!\!,pos(j))^{\top}
}{
    \sqrt{d_h}
}
\nonumber\\
&+ M_{ij}+b_{ij}^{(\ell,h)}.
\label{eq:topological_attention}
\end{align}where $d_h$ denotes the feature dimension of each attention head. We apply RoPE~\citep{su2021roformer,liu2026ml} with CPE as input to mitigate the effects of input-order variation. $\emph{Q}_i$ denotes the token representation of the $i$-th input embedding vector and serves as the query vector, while $\emph{K}_j$ denotes the token representation of the $j$-th input embedding vector and serves as the key vector. Specifically, $i, j\in\mathcal{Z}\cup\mathcal{X}$, where $\mathcal{X} = \bigcup_{c \in \mathcal{C}}\mathcal{X}^c$.




%% file: sections/05_Experiments.tex
\newenvironment{tightitemize}
{%
  \begin{list}{\textbullet}{%
    \setlength{\leftmargin}{1.15em}%
    \setlength{\labelwidth}{0.45em}%
    \setlength{\labelsep}{0.35em}%
    \setlength{\itemindent}{0pt}%
    \setlength{\listparindent}{0pt}%
    \setlength{\rightmargin}{0pt}%
    \setlength{\topsep}{1pt}%
    \setlength{\partopsep}{0pt}%
    \setlength{\parsep}{0pt}%
    \setlength{\itemsep}{0pt}%
  }%
}
{%
  \end{list}%
}

\section{Experiments}
\label{sec:experiments}
We conduct comprehensive experiments on \dname to evaluate different LLMs and address the following research questions: RQ1: How well do state-of-the-art LLMs perform on \dname, particularly in maintaining factual consistency under input-order variations of knowledge contexts? RQ2: How does the proposed \pname perform compared with four baselines across different LLM backbones? RQ3: How do different components, including CPE, LBAM, and TKB, affect the performance of \pname? 



\subsection{Experimental Setup}
\label{sec:experimental_setup}


\paragraph{Evaluated LLMs.}
To obtain a comprehensive assessment of current LLMs on \dname, we evaluate both open-source and closed-source models. The open-source models include Qwen3-8B, Qwen3-30B-A3B, Gemma-3-12B, MiniMax-M2.7, Kimi-K2.5, DeepSeek-V3.2, and DeepSeek-V4-Flash~\citep{yang2025qwen3,gemmateam2025gemma3,minimax2026m2,kimiteam2026kimik25,deepseekai2025v32,deepseekai2026v4}. The closed-source models include Gemini-2.5-Flash-Lite, Gemini-2.5-Flash~\citep{geminiteam2025gemini25}, GPT-4.1 mini, GPT-5, Grok-4.3, o4-mini, and o3.

For the backbone of \pname, we use four different LLMs: Qwen3-8B, GLM-4-9B-Chat~\citep{teamglm2024chatglm}, Llama-3.1-8B-Instruct~\citep{grattafiori2024llama3}, and Mistral-7B-Instruct-v0.3~\citep{jiang2023mistral}.

\paragraph{Baselines.}
We compare \pname with the following training-free (TF) and LoRA-based (LB) baselines: \textbf{Direct QA}, \textbf{ReAct}~\citep{yao2022react}, \textbf{CoT-D}~\citep{wang2024chain}, \textbf{TXH}~\citep{zhao2020transformer}, and \textbf{PMFT}~\citep{huang2025masking}. Detailed descriptions of these baselines and their corresponding prompts are provided in \textbf{Appendix~\ref{app:baseline_details}}.





\paragraph{Evaluation Metrics}
As described in Section~\ref{sec:dataset}, we adopt three evaluation metrics: \textbf{Exact Match (EM)} for measuring answer correctness against the gold answers, \textbf{Reasoning-Chain Accuracy (RCA)} for evaluating reasoning-chain correctness, and \textbf{Order Standard Deviation (O. Std.)} for assessing robustness to input-order variations. For EM and RCA, we report the average (\textbf{Avg. EM} and \textbf{Avg. RCA}), best (\textbf{Best EM} and \textbf{Best RCA}), and worst (\textbf{Worst EM} and \textbf{Worst RCA}) scores.

\input{tables/Order_preformance}

\paragraph{Implementation Details}
All state-of-the-art LLMs are evaluated on the test split of \dname. LoRA-based methods are trained on the training split and evaluated on the test split. All experiments are repeated three times with different random seeds, and we report the average results. Additional implementation details are provided in \textbf{Appendix~\ref{app:implementation_details}}.

\input{tables/method_evaluation}

\subsection{Performance Analysis of SOTA LLMs (RQ1)}
\label{sec:benchmark_evaluation}
To assess the ability of SOTA LLMs to maintain factuality consistency across heterogeneous knowledge contexts, including tables (T), text (P), and KG (G), we evaluate 14 open- and closed-source LLMs on \dname. We further report results under different input-order permutations, such as Table--Text--KG, denoted as T-P-G, to examine robustness to input-order variations.
Table~\ref{tab:order_wise_performance} reports EM and RCA under different input orders, along with their corresponding O. Std. values. 

The results reveal that maintaining factuality consistency across heterogeneous knowledge contexts remains a significant challenge for all SOTA LLMs. Specifically, DeepSeek-V4-Flash, the most advanced LLM, achieves average EM and RCA scores of 85.18\% and 65.14\%, respectively. Another notable baseline, GPT-5, also shows unsatisfactory performance in this regard, achieving average EM and RCA scores of 76.80\% and 52.52\%, respectively. Notably, all evaluated LLMs exhibit consistently lower RCA scores than EM scores, indicating that correct final answers are not necessarily supported by multi-hop reasoning chains grounded in the provided knowledge contexts. This finding indicates that current SOTA LLMs sometimes fail to understand heterogeneous knowledge contexts and conduct effective knowledge-grounded multi-hop reasoning. Detailed examples and inference prompts for the SOTA LLMs are provided in the \textbf{Appendix~\ref{app:sota_llm_example}}. Furthermore, the results indicate that all evaluated LLMs are sensitive to variations in the input order of heterogeneous knowledge contexts. For example, Gemma-3-12B exhibits O. Std. values of 1.42 for EM and 1.59 for RCA, while Grok-4.3 exhibits O. Std. values of 0.67 and 3.48, respectively.


\subsection{Comparison of \pname with Baselines (RQ2)}
\label{sec:performance_comparison}
To evaluate the effectiveness of \pname in maintaining factual consistency while enhancing robustness to input-order variations, we compare the proposed framework with four baseline methods using four different LLM backbones. Specifically, the baselines are grouped into training-free (TF) and LoRA-based (LB) methods. Table~\ref{tab:method_comparison} presents the main comparison results,
while the complete results across all four backbones are provided in \textbf{Appendix~\ref{app:additional_performance_comparison}}. We report the best, worst, and order-wise standard deviation values for both EM and RCA. The best result for each backbone and metric is highlighted in bold.

Our framework outperforms all baselines across different LLM backbones, achieving average improvements of 2.15\% in Avg. EM and 4.29\% in Avg. RCA, respectively. Table~\ref{tab:method_comparison} demonstrates that \pname preserves factuality consistency more effectively than other SOTA baselines when generating responses based on heterogeneous knowledge contexts. This improvement is mainly attributed to TKB, which enables \pname to encode topological relations among heterogeneous knowledge contexts. Meanwhile, compared with all baselines across four LLM backbones, \pname reduces O. Std. by 0.04--3.71 for EM and 0.01--3.01 for RCA, respectively, demonstrating its effectiveness in improving performance robustness across different LLM backbones.

\subsection{Ablation Study (RQ3)}
\label{sec:ablation_study}

\begin{figure}[htbp]
    \vspace{-2mm}
    \centering
    \includegraphics[width=\columnwidth]{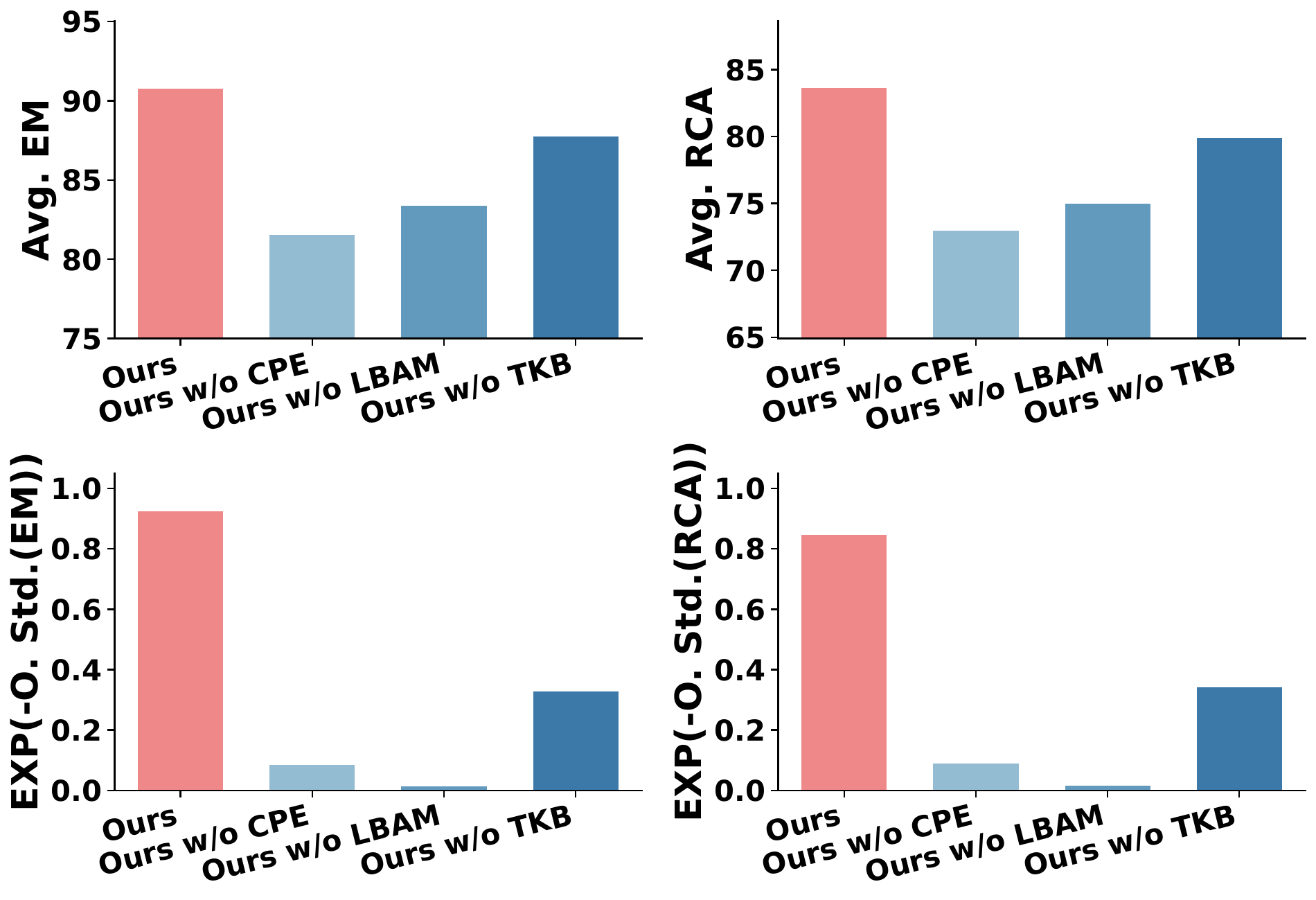}
    \vspace{-3mm}
    \caption{Ablation results of {\pname} and its variants on Qwen3-8B.}
    \label{fig:ablation}
    \vspace{-5mm}
\end{figure}

To validate the effectiveness of the key components in \pname, we
construct three ablated variants by replacing CPE with standard global
position IDs, replacing LBAM with standard causal attention, and
disabling TKB so that no topology bias is introduced. 
Figure~\ref{fig:ablation} presents the overall ablation results, while
detailed numerical results are provided in
Table~\ref{tab:ablation_results} of
\textbf{Appendix~\ref{app:additional_ablation}}.

The results show that all three components contribute to the effectiveness of \pname. Compared with all ablated variants across the four LLM backbones, the complete framework achieves average relative improvements of 9.83\% in Avg. EM and 13.18\% in Avg. RCA, demonstrating its effectiveness in maintaining factual consistency across heterogeneous knowledge contexts. Moreover, \pname reduces O. Std. by 0.69--4.51 for EM and 0.68--4.27 for RCA, confirming its effectiveness in improving robustness to input-order variations. Specifically, CPE encodes context-wise positional information, LBAM reduces cross-context interference under different input orders, and TKB captures topological dependencies among heterogeneous knowledge contexts to support reliable multi-hop reasoning. A sensitivity analysis of the number of latent tokens is provided in \textbf{Appendix~\ref{app:sensitivity}}.

\section{Conclusion}

In this work, we introduced \dname, a benchmark for evaluating factual consistency and input-order robustness in multi-hop grounded reasoning over tables, textual passages, and knowledge graphs. Evaluations of 14 open- and closed-source LLMs reveal that current models remain limited in reasoning-chain accuracy and sensitive to variations in knowledge-context order. To address these limitations, we proposed \pname, an LLM-agnostic framework that models cross-context topological relations through knowledge-specific latent vectors. Experiments across four LLM backbones demonstrate that \pname consistently improves answer and reasoning-chain accuracy while enhancing robustness to input-order variations. Future work will extend \dname and \pname to multimodal knowledge contexts and more diverse real-world grounded-reasoning scenarios.

%% file: tables/Order_preformance.tex
\begin{table*}[t]
    \centering
    \vspace{-2mm}
    \small
    \setlength{\tabcolsep}{3.4pt}
    \renewcommand{\arraystretch}{1.10}

    \resizebox{\textwidth}{!}{%
    \begin{tabular}{
        l
        ccccccc
        @{\hspace{7pt}}
        ccccccc
    }
        \toprule

        \multirow{2}{*}{\textbf{Model}}
        & \multicolumn{7}{c}{\textbf{Exact Match (EM) $\uparrow$}}
        & \multicolumn{7}{c}{\textbf{Reasoning-Chain Accuracy (RCA) $\uparrow$}} \\

        \cmidrule(lr){2-8}
        \cmidrule(lr){9-15}

        & \textbf{T-P-G}
        & \textbf{T-G-P}
        & \textbf{P-T-G}
        & \textbf{P-G-T}
        & \textbf{G-P-T}
        & \textbf{G-T-P}
        & \textbf{O. Std. $\downarrow$}
        & \textbf{T-P-G}
        & \textbf{T-G-P}
        & \textbf{P-T-G}
        & \textbf{P-G-T}
        & \textbf{G-P-T}
        & \textbf{G-T-P}
        & \textbf{O. Std. $\downarrow$} \\

        \midrule

        \multicolumn{15}{c}{\textbf{\textit{Open-Source Models}}} \\
        \cmidrule(lr){1-15}

        Qwen3-8B
        & 52.35 & 46.20 & 52.18 & 54.36 & 44.63 & 45.96
        & 3.78
        & 20.14 & 18.87 & 19.55 & 22.48 & 22.06 & 19.65
        & 1.34 \\

        Qwen3-30B
        & 64.90 & 64.41 & 65.29 & 65.36 & 64.97 & 65.28
        & \textbf{0.33}
        & 43.31 & 44.42 & 43.56 & 47.81 & 48.03 & 44.48
        & 1.92 \\

        Gemma-3-12B
        & 62.33 & 59.89 & 61.85 & 63.89 & 61.33 & 59.78
        & 1.42
        & 19.95 & 19.49 & 20.49 & 21.71 & 21.26 & 16.81
        & 1.59 \\

        MiniMax-M2.7
        & 72.42 & 72.49 & 72.59 & 71.28 & 72.33 & 72.57
        & 0.46
        & 55.34 & 54.33 & 53.66 & 55.79 & 56.69 & 55.86
        & 1.01 \\

        Kimi-K2.5
        & 79.31 & 77.96 & 80.09 & 78.88 & 79.33 & 80.38
        & 0.79
        & 64.04 & 65.04 & 64.93 & 65.24 & 67.70 & 68.60
        & 1.64 \\

        DeepSeek-V3.2
        & 83.05 & 82.24 & 82.25 & 82.60 & 82.98 & 83.42
        & 0.43
        & 69.20 & 71.94 & 71.24 & 72.55 & 75.05 & 70.63
        & 1.81 \\

        DeepSeek-V4-Flash
        & \textbf{84.66}
        & 84.87
        & 83.90
        & \textbf{85.44}
        & \textbf{86.64}
        & \textbf{85.57}
        & 0.85
        & 65.26
        & 64.21
        & 62.34
        & 64.85
        & 67.59
        & 66.59
        & 1.68 \\

        \midrule

        \multicolumn{15}{c}{\textbf{\textit{Closed-Source Models}}} \\
        \cmidrule(lr){1-15}

        Gemini-2.5-Flash-Lite
        & 69.93 & 67.67 & 70.52 & 68.19 & 71.52 & 71.56
        & 1.51
        & 37.25 & 39.68 & 37.98 & 37.55 & 46.26 & 41.87
        & 3.17 \\

        Gemini-2.5-Flash
        & 84.20
        & 84.68
        & 83.30
        & 83.64
        & 85.55
        & 85.48
        & 0.85
        & 77.93
        & 79.62
        & 75.68
        & 77.38
        & \textbf{80.57}
        & \textbf{79.92}
        & 1.69 \\

        GPT-4.1-Mini
        & 71.44 & 71.00 & 70.44 & 69.22 & 74.30 & 74.97
        & 2.06
        & 29.59 & 34.38 & 27.94 & 29.37 & 38.52 & 35.10
        & 3.78 \\

        GPT-5
        & 75.38 & 77.13 & 76.38 & 77.38 & 76.75 & 77.75
        & 0.77
        & 51.52 & 53.30 & 52.43 & 50.76 & 54.89 & 52.23
        & 1.32 \\

        Grok-4.3
        & 79.22 & 78.31 & 78.89 & 77.78 & 79.87 & 78.56
        & 0.67
        & 32.61 & 33.00 & 31.03 & 33.17 & 41.26 & 37.27
        & 3.48 \\

        o4-mini
        & 82.50 & 80.88 & 81.25 & 80.25 & 84.25 & 84.25
        & 1.58
        & 68.18 & 71.18 & 71.75 & 69.25 & 72.48 & 72.68
        & 1.66 \\

        o3
        & 84.50
        & \textbf{86.25}
        & \textbf{85.63}
        & 84.75
        & 83.88
        & 85.13
        & 0.77
        & \textbf{79.05}
        & \textbf{80.01}
        & \textbf{79.87}
        & \textbf{80.63}
        & 80.09
        & 79.61
        & \textbf{0.48} \\

        \bottomrule
    \end{tabular}%
    }
\vspace{-3mm}
\caption{
Performance of SOTA LLMs on \dname.
}

    \label{tab:order_wise_performance}
\end{table*}

%% file: tables/method_evaluation.tex

\definecolor{oursblue}{RGB}{226,239,249}

\begin{table*}[t]
    \centering
    \small
    \setlength{\tabcolsep}{4.2pt}
    \renewcommand{\arraystretch}{1.10}

    \resizebox{\textwidth}{!}{%
    \begin{tabular}{lllcccccccc}
        \toprule

        \textbf{Backbone}
        & \textbf{Setting}
        & \textbf{Method}
        & \textbf{Avg. EM $\uparrow$}
        & \textbf{Best EM $\uparrow$}
        & \textbf{Worst EM $\uparrow$}
        & \textbf{Avg. RCA $\uparrow$}
        & \textbf{Best RCA $\uparrow$}
        & \textbf{Worst RCA $\uparrow$}
        & \textbf{O. Std. (EM) $\downarrow$}
        & \textbf{O. Std. (RCA) $\downarrow$} \\

        \midrule

        \multirow{6}{*}{Qwen3-8B}
        & \multirow{3}{*}{TF}
        & Direct QA
        & 49.28
        & 54.36
        & 44.63
        & 20.46
        & 22.48
        & 18.87
        & 3.78
        & 1.34 \\

        & & ReAct
        & 53.70
        & 58.10
        & 49.40
        & 27.22
        & 29.50
        & 25.10
        & 3.21
        & 1.58 \\

        & & CoT-D
        & 57.83
        & 61.30
        & 54.60
        & 34.03
        & 36.20
        & 31.80
        & 2.55
        & 1.52 \\

        \cmidrule(lr){2-11}

        & \multirow{3}{*}{LB}
        & TXH
        & 87.75
        & 87.83
        & 87.50
        & 78.94
        & 79.17
        & 78.67
        & 0.13
        & 0.18 \\

        & & PMFT
        & 87.89
        & \textbf{92.33}
        & 83.67
        & 79.38
        & 83.50
        & 75.67
        & 3.79
        & 3.18 \\

        \rowcolor{oursblue}
        & & \textbf{Ours}
        & \textbf{90.72}
        & 90.83
        & \textbf{90.67}
        & \textbf{83.61}
        & \textbf{83.83}
        & \textbf{83.33}
        & \textbf{0.08}
        & \textbf{0.17} \\

        \bottomrule
    \end{tabular}%
    }
\vspace{-2mm}
\caption{
Performance comparison of \pname and baseline methods on Qwen3-8B.
}
    \label{tab:method_comparison}
    \vspace{-5mm}
\end{table*}

%% file: sections/Limitations.tex
\section{Limitations}

The limitations of our work are as follows:
1) \dname is constructed from English Wikipedia tables, textual passages,
and Wikidata KG subgraphs. Its generalizability to other languages,
domains, and knowledge sources remains to be further investigated.
2) \pname requires access to the internal attention mechanism and
positional encoding of the backbone model, and therefore cannot be
directly applied to closed-source LLMs accessible only through APIs.

%% file: sections/Appendix.tex
\clearpage

\appendix




In this appendix, we provide supplementary details and
additional analyses that are omitted from the main paper due
to space constraints. In particular, the appendix contains
the following:

\begin{itemize}
    \item \hyperref[app:benchmark_details]{
        Additional Benchmark Details
    }
    \item \hyperref[app:additional_experiments]{
        Additional Experimental Results
    }
    \item \hyperref[app:baseline_details]{
        Baseline Details
    }
    \item \hyperref[app:implementation_details]{
        Implementation Details
    }
    \item \hyperref[app:Prompt]{
        Prompt Templates
    }
\end{itemize}

\input{appendix/A.Benchmark_details.tex}
\input{appendix/B.Additional_Experimental_Results.tex}
\input{appendix/C.Baseline_Details.tex}
\input{appendix/D.Implementation_Details.tex}
\input{appendix/E.Prompts_and_Evaluation_protocol.tex}

%% file: appendix/A.Benchmark_details.tex
\section{Additional Benchmark Details}
\label{app:benchmark_details}

This section provides three supplementary views of \dname: a complete
counterfactual context example, the additional construction details, and a comparative case study of representative SOTA LLMs.


\subsection{Counterfactual Context Example}
\label{app:dataset_example}

\begin{figure*}[!t]
    \centering

    \includegraphics[
        width=0.98\textwidth,
        height=0.4\textheight,
        keepaspectratio
    ]{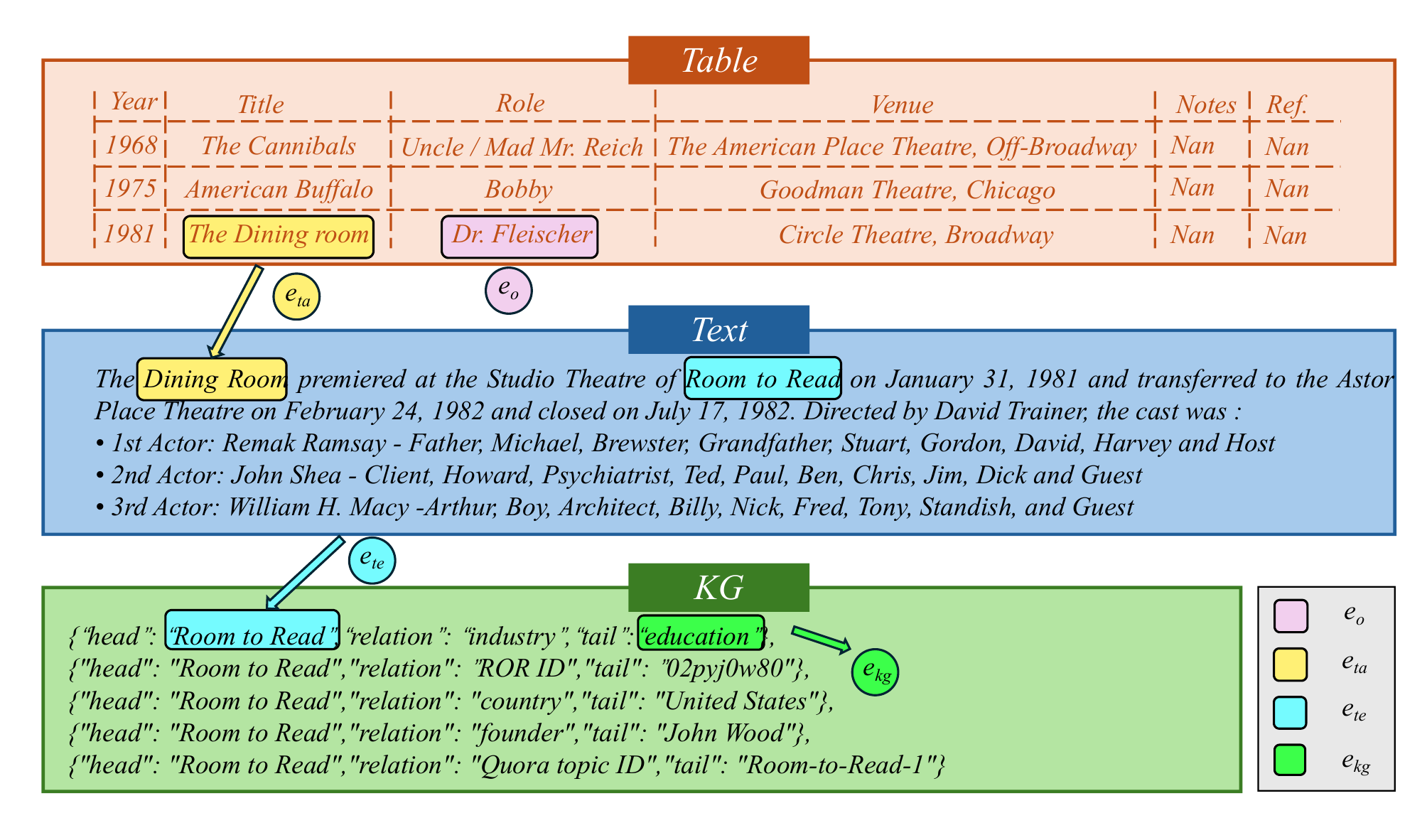}

    \caption{
    An example of a counterfactual context in \dname. The highlighted
entities form the corresponding counterfactual reasoning chain across
the table, textual passage, and KG subgraph.
    }
    \label{fig:dataset_sample}
\end{figure*}

Figure~\ref{fig:dataset_sample} presents a counterfactual context
consisting of a table, a textual passage, and a KG subgraph. The
highlighted entities $e_o$, $e_{ta}$, $e_{te}$, and $e_{kg}$ connect
the three heterogeneous knowledge contexts and form the annotated
counterfactual reasoning chain used for question generation.

\subsection{Additional Construction Details}
\label{app:additional_construction}

\subsubsection{QA Pair Generation}
\begin{figure*}[!t]
    \centering

    \includegraphics[
        width=0.95\textwidth,
        height=0.5\textheight,
        keepaspectratio,
    ]{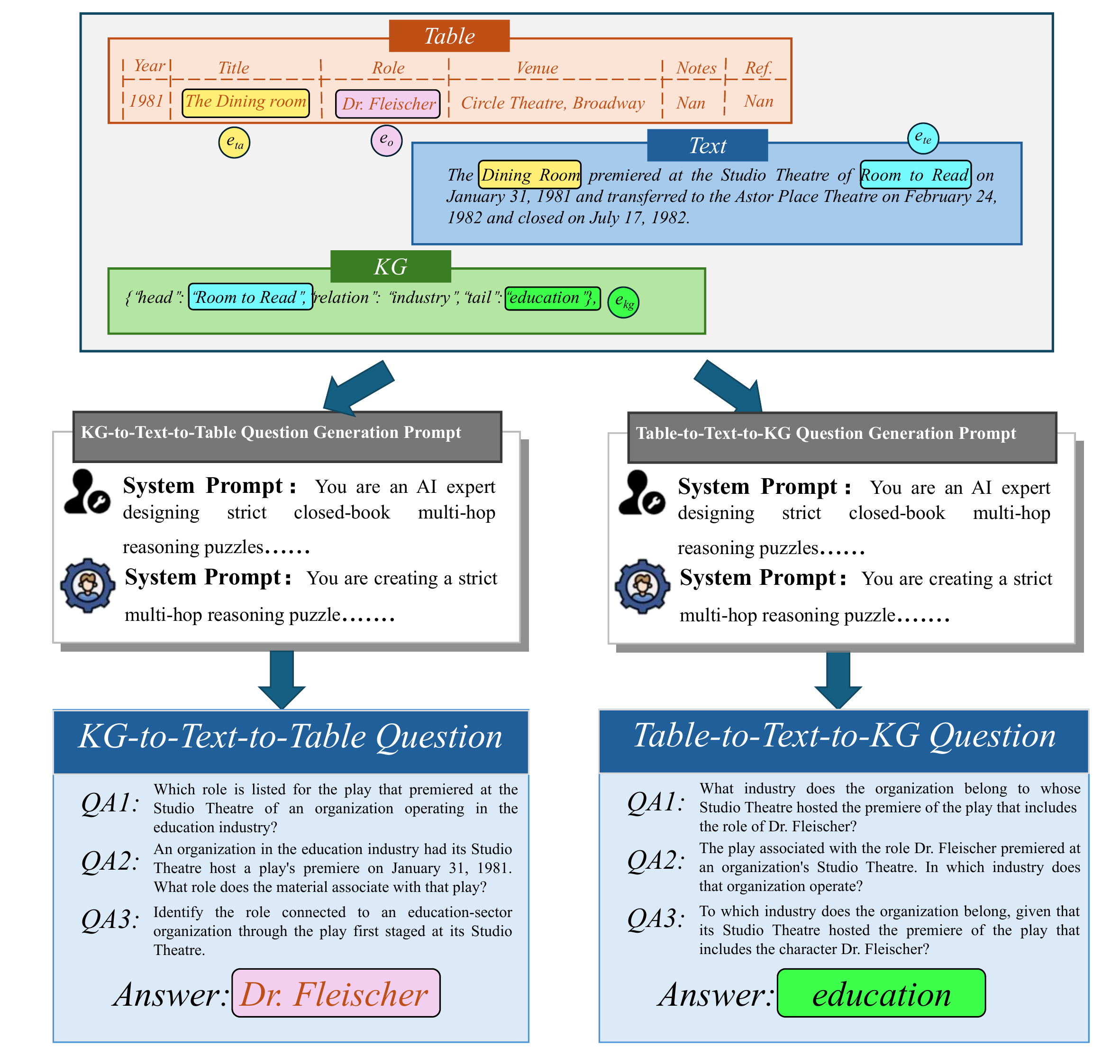}

    \caption{
    Bidirectional question generation from a counterfactual reasoning
    chain in \dname.
    }
    \label{fig:generation_image}
\end{figure*}

To mitigate potential performance degradation arising from limited question diversity\citep{chia-etal-2025-longdoc}, we employ three different LLMs to independently generate candidate questions for each counterfactual reasoning chain $\tilde{\mathcal{C}} = \langle\tilde{T}, \tilde{P}, \tilde{G}\rangle(e_o, e_{ta}, e_{te}, e_{kg})$. We then randomly select one of the three candidate questions as the final question corresponding to the counterfactual reasoning chain. An detailed example of different question forms corresponding to the same counterfactual reasoning chain is provided in ~\ref{fig:generation_image}. To further improve the diversity of our benchmark data, we generate questions for each counterfactual reasoning chain along two reasoning directions: Table-to-Text-to-KG and KG-to-Text-to-Table. These two directions correspond to different final answer entities, namely $e_{kg}$ and $e_o$, respectively. Part 2 of Figure~\ref{fig:orcm_overview} illustrates the questions generated from two reasoning directions based on a single counterfactual reasoning chain, with the question-generation prompt provided in ~\ref{fig:qa_generation_prompt}.


\subsubsection{Automatic Verification and Manual Review}
\label{sec:quality_control}
To ensure that entity replacement is reasonable and that the generated QA pairs are correct, we manually inspect the generated QA pairs and counterfactual reasoning chains from three perspectives: 1) Chain Consistency, which assesses the logical coherence of the constructed reasoning chain across tables, text, and KGs; 2) Entity Alignment, which verifies that the replaced entities are correctly aligned across heterogeneous sources and belong to the intended semantic categories; and 3) Final Answer Validation, which checks whether the final answer can be derived from the constructed counterfactual reasoning chain.



\subsection{Responses from State-of-the-Art LLMs}
\label{app:sota_llm_example}

\begin{figure*}[!t]
    \centering

    \includegraphics[
        width=0.95\textwidth,
        height=0.8\textheight,
        keepaspectratio,
    ]{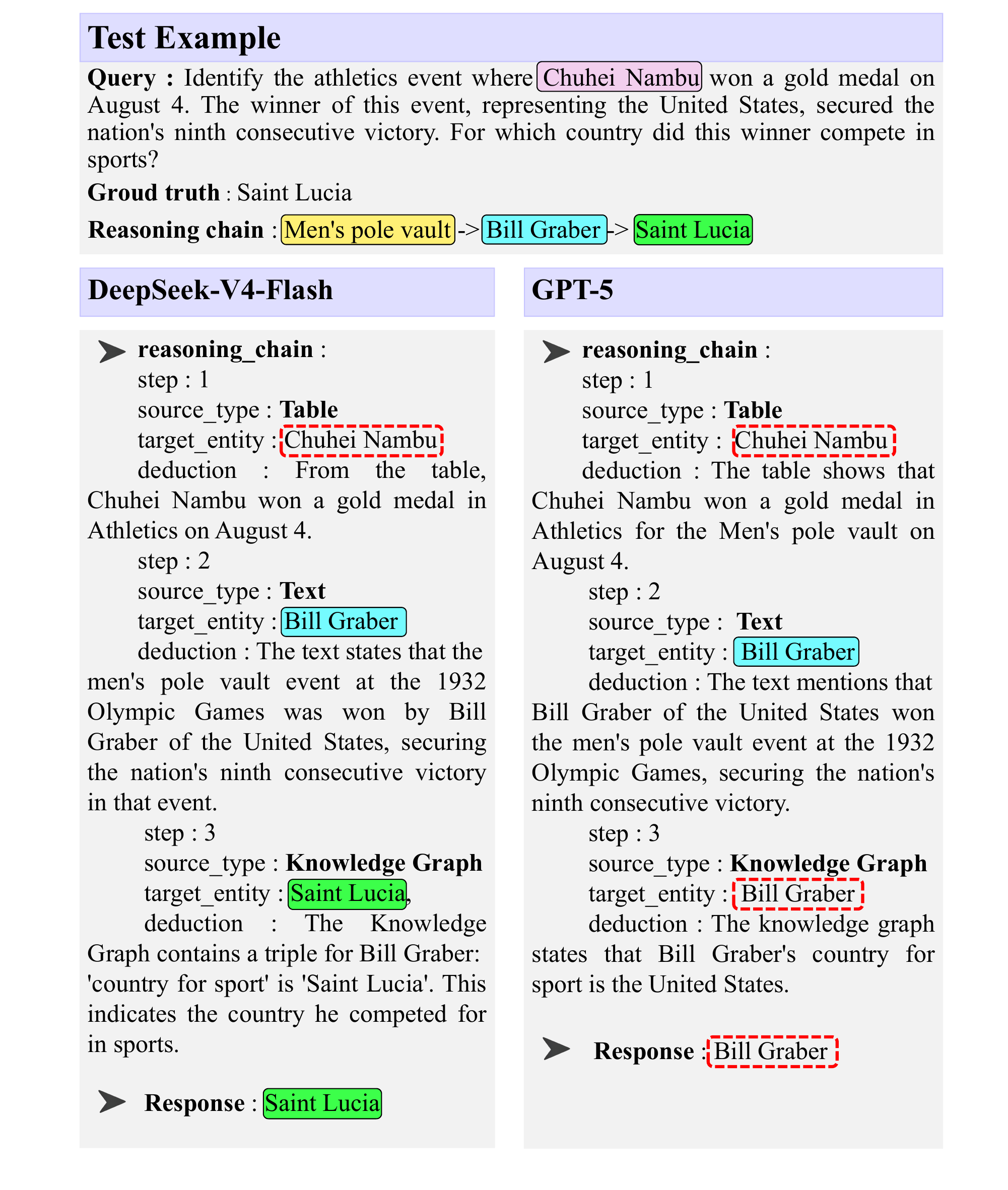}

 \caption{
Comparative case study of DeepSeek-V4-Flash and GPT-5 on the same
\dname question. The text highlighted in pink denotes $e_o$, while
the text highlighted in yellow, blue, and green denotes $e_{ta}$,
the target entity in the table, $e_{te}$, the target entity in the
textual passage, and $e_{kg}$, the target entity in the KG and the
final answer, respectively. Red dashed boxes indicate incorrectly
predicted intermediate entities or final answers. As shown in the
left panel, DeepSeek-V4-Flash produces the correct final answer despite
an error in its intermediate reasoning chain, whereas GPT-5 in the
right panel makes incorrect predictions in both the reasoning chain
and the final answer.
}
    \label{fig:sota_llm_case_study}
\end{figure*}

Figure~\ref{fig:sota_llm_case_study} compares the responses of
DeepSeek-V4-Flash and GPT-5 to the same \dname question under identical
heterogeneous knowledge contexts. As shown in the left panel,
DeepSeek-V4-Flash derives an incorrect intermediate entity but still
produces the correct final answer. By contrast, GPT-5 in the right panel
makes errors in both its intermediate reasoning chain and final answer.
This example demonstrates that final-answer correctness does not
necessarily indicate a fully correct reasoning process.

%% file: appendix/B.Additional_Experimental_Results.tex
\section{Additional Experimental Results}
\label{app:additional_experiments}

\input{appendix/tables/method_evaluation}

This section presents supplementary results for the ablation study and
sensitivity analysis. We provide complete ablation results across four
LLM backbones to examine the contribution and generalizability of each
component in \pname, and further investigate how the number of latent
vectors affects performance and robustness to input-order variations.

\subsection{Additional Performance Comparison}
\label{app:additional_performance_comparison}

Table~\ref{tab:method} presents the complete comparison
of \pname with the training-free and LoRA-based baselines across four
LLM backbones. The results include the average, best, worst, and O. Std.
values of EM and RCA over the six input-order permutations. Overall,
\pname achieves the best Avg. RCA across all four backbones and the best
Avg. EM on three of them, while consistently obtaining substantially
lower O. Std. values. These results further confirm that \pname improves
both factual consistency and robustness to input-order variations across
different model architectures.

\subsection{Ablation Results}
\label{app:additional_ablation}

The main paper presents the ablation results on Qwen3-8B. To further
examine whether the contributions of the proposed components generalize
across different model architectures, we conduct the same ablation study
on three additional backbones: GLM-4-9B-Chat,
Llama-3.1-8B-Instruct, and Mistral-7B-Instruct-v0.3. Complete numerical
results across all four backbones are reported in
Table~\ref{tab:ablation_results}.

\input{tables/ablation_table}

\begin{figure*}[p]
    \centering
    \captionsetup[subfigure]{
        font=small,
        justification=centering,
        skip=1pt
    }

    {\renewcommand{\thesubfigure}{Q-\alph{subfigure}}
    \setcounter{subfigure}{0}

    \begin{subfigure}[t]{0.238\textwidth}
        \centering
        \includegraphics[width=\linewidth]
        {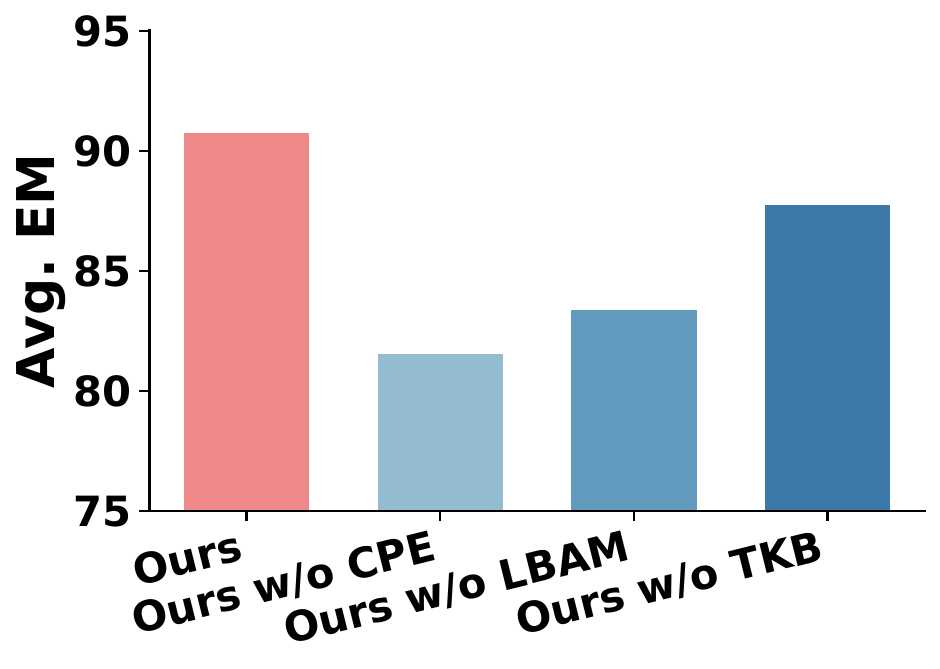}
        \caption{}
        \label{fig:ablation_qwen_avg_em}
    \end{subfigure}
    \hfill
    \begin{subfigure}[t]{0.238\textwidth}
        \centering
        \includegraphics[width=\linewidth]
        {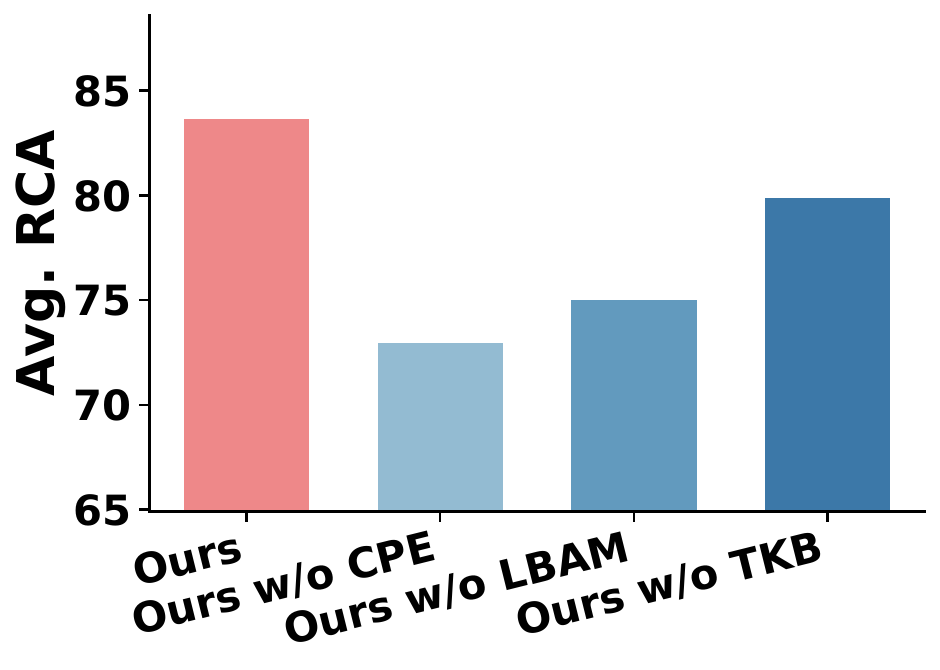}
        \caption{}
        \label{fig:ablation_qwen_avg_rca}
    \end{subfigure}
    \hfill
    \begin{subfigure}[t]{0.238\textwidth}
        \centering
        \includegraphics[width=\linewidth]
        {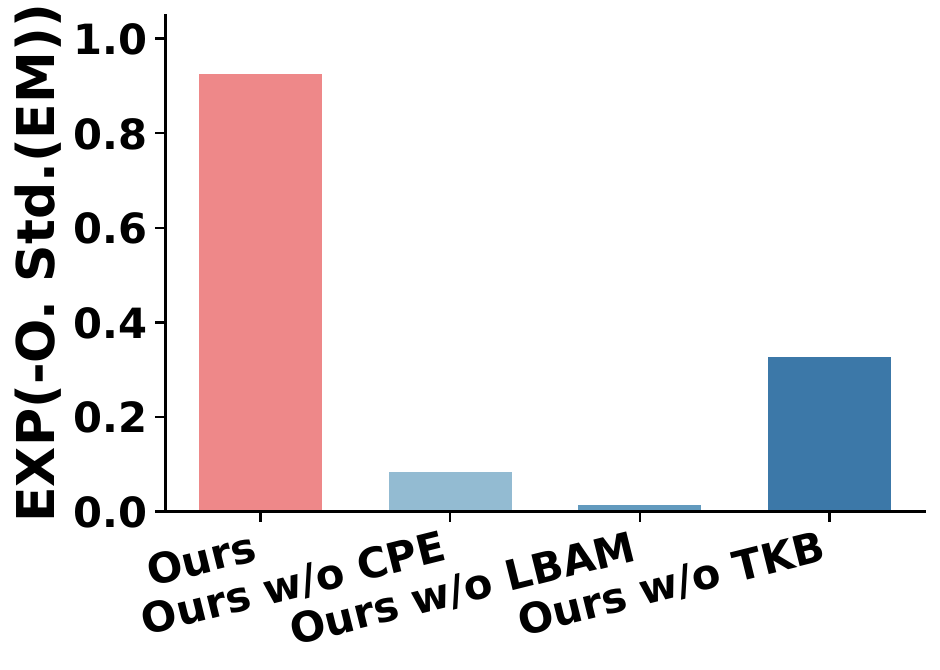}
        \caption{}
        \label{fig:ablation_qwen_ostd_em}
    \end{subfigure}
    \hfill
    \begin{subfigure}[t]{0.238\textwidth}
        \centering
        \includegraphics[width=\linewidth]
        {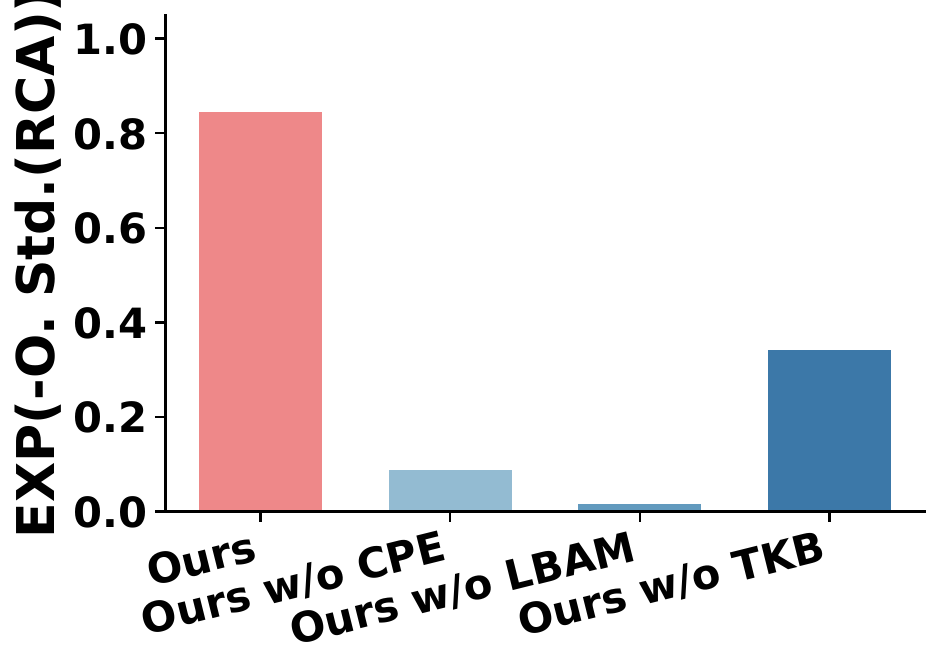}
        \caption{}
        \label{fig:ablation_qwen_ostd_rca}
    \end{subfigure}
    }

    \vspace{4pt}

    {\renewcommand{\thesubfigure}{G-\alph{subfigure}}
    \setcounter{subfigure}{0}

    \begin{subfigure}[t]{0.238\textwidth}
        \centering
        \includegraphics[width=\linewidth]
        {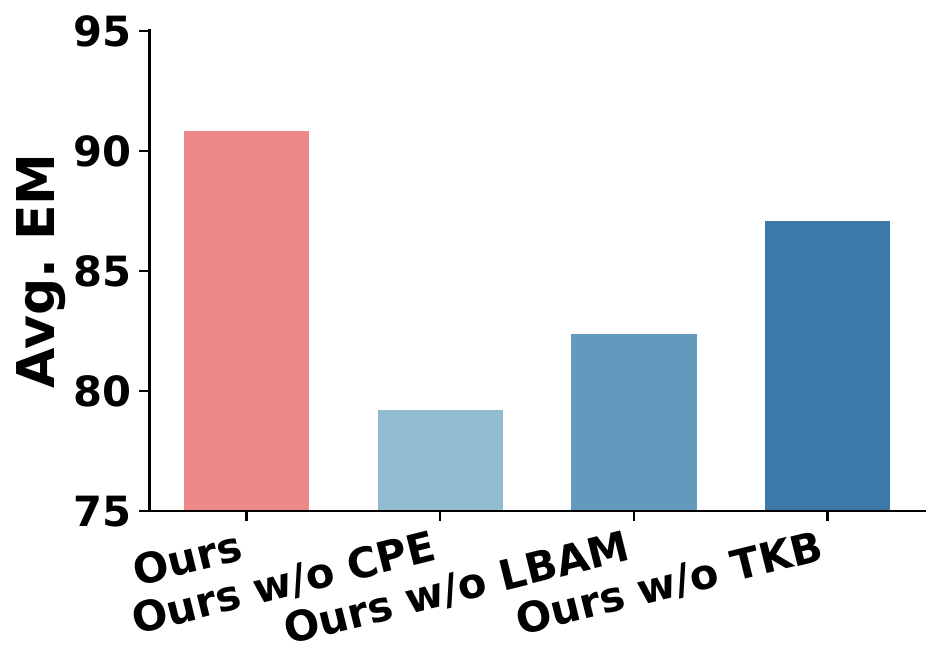}
        \caption{}
        \label{fig:ablation_glm_avg_em}
    \end{subfigure}
    \hfill
    \begin{subfigure}[t]{0.238\textwidth}
        \centering
        \includegraphics[width=\linewidth]
        {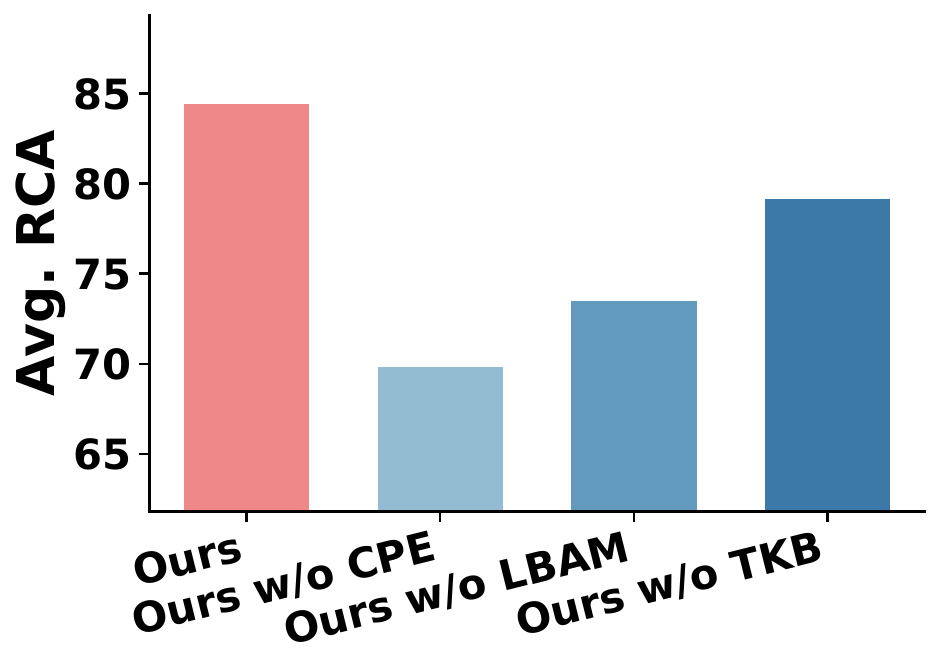}
        \caption{}
        \label{fig:ablation_glm_avg_rca}
    \end{subfigure}
    \hfill
    \begin{subfigure}[t]{0.238\textwidth}
        \centering
        \includegraphics[width=\linewidth]
        {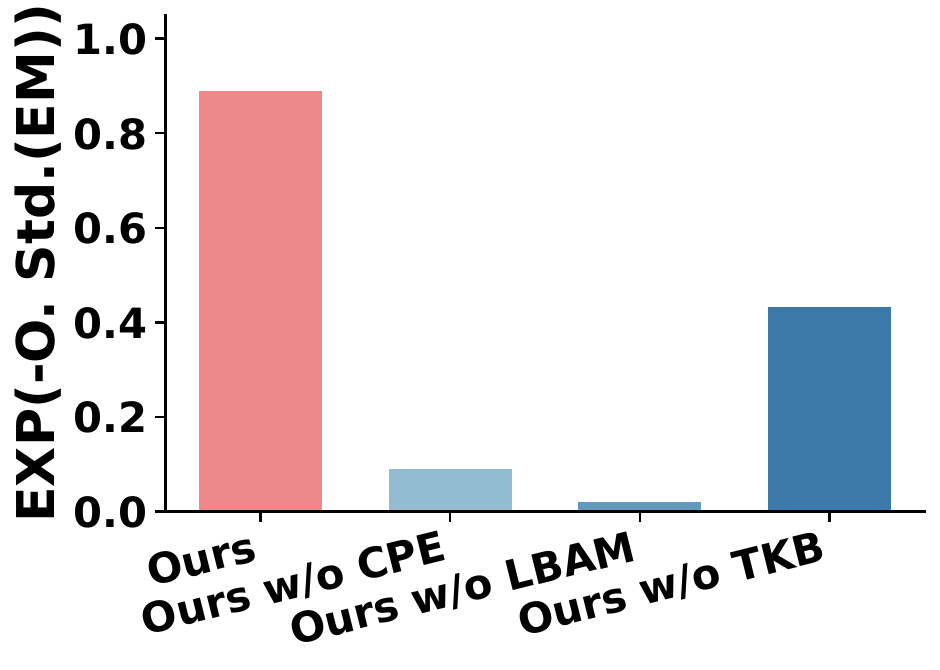}
        \caption{}
        \label{fig:ablation_glm_ostd_em}
    \end{subfigure}
    \hfill
    \begin{subfigure}[t]{0.238\textwidth}
        \centering
        \includegraphics[width=\linewidth]
        {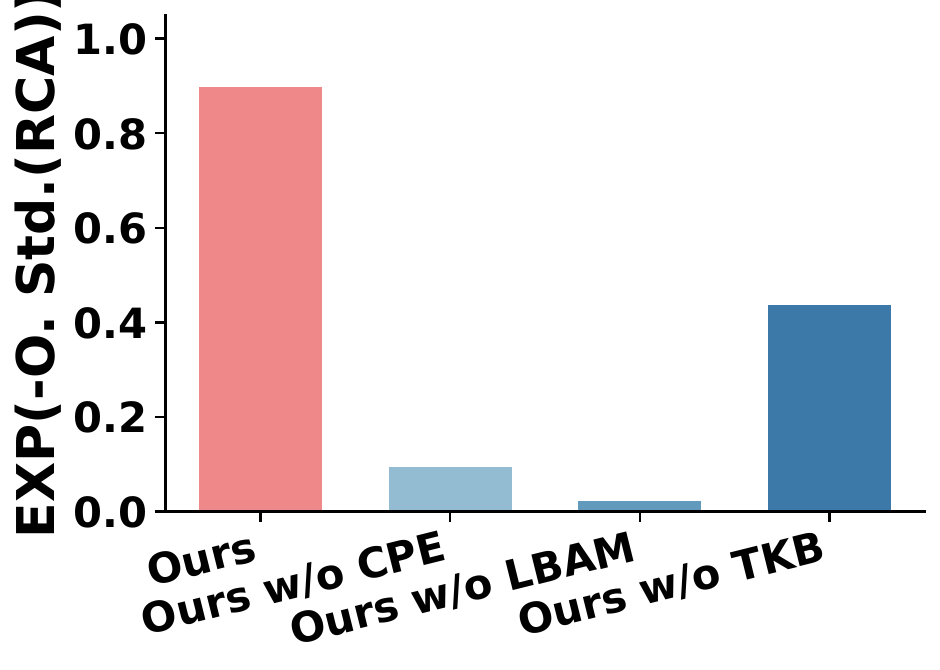}
        \caption{}
        \label{fig:ablation_glm_ostd_rca}
    \end{subfigure}
    }

    \vspace{4pt}

    {\renewcommand{\thesubfigure}{L-\alph{subfigure}}
    \setcounter{subfigure}{0}

    \begin{subfigure}[t]{0.238\textwidth}
        \centering
        \includegraphics[width=\linewidth]
        {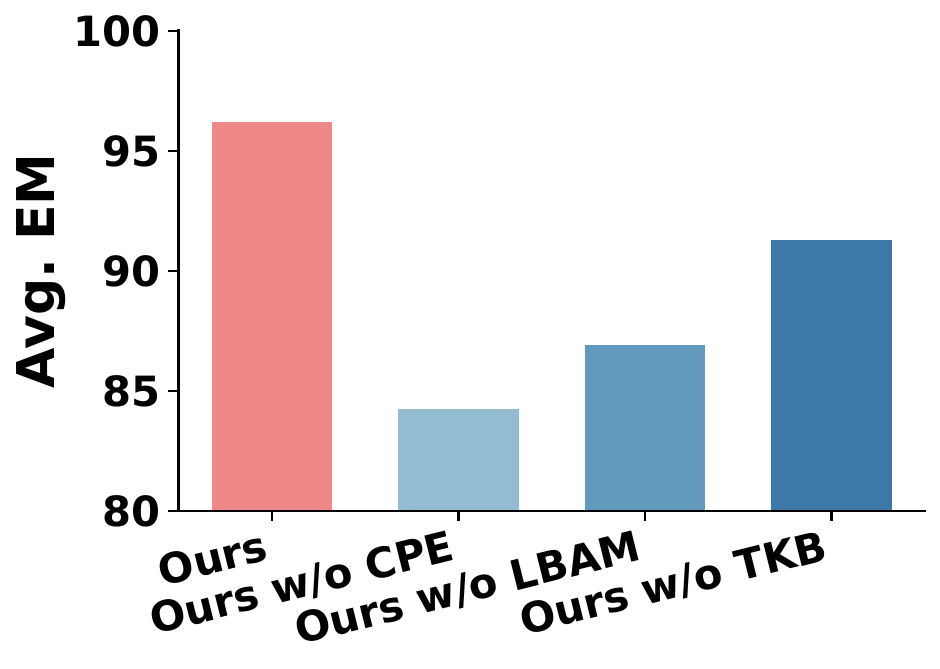}
        \caption{}
        \label{fig:ablation_llama_avg_em}
    \end{subfigure}
    \hfill
    \begin{subfigure}[t]{0.238\textwidth}
        \centering
        \includegraphics[width=\linewidth]
        {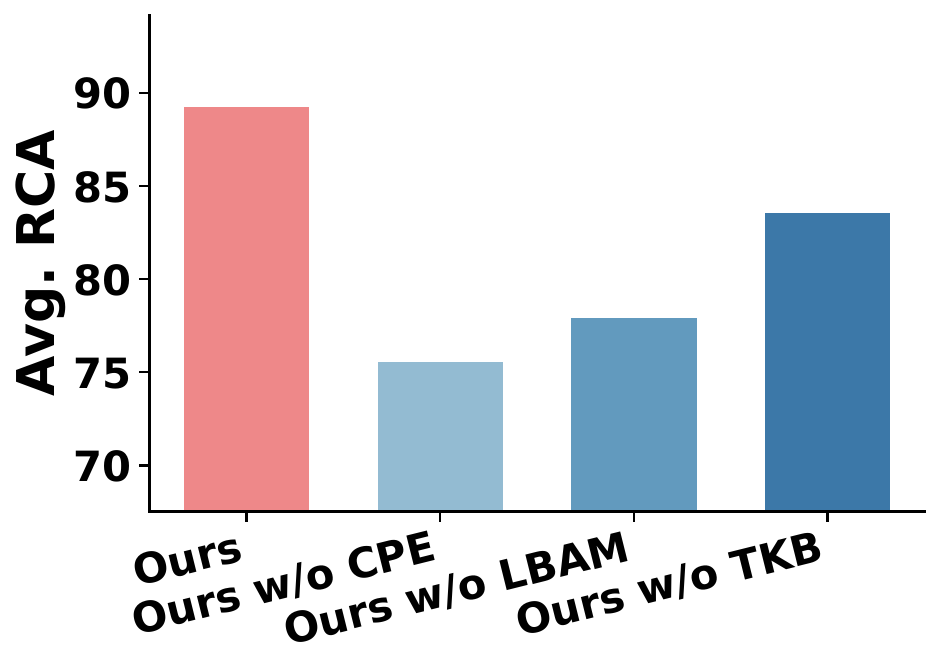}
        \caption{}
        \label{fig:ablation_llama_avg_rca}
    \end{subfigure}
    \hfill
    \begin{subfigure}[t]{0.238\textwidth}
        \centering
        \includegraphics[width=\linewidth]
        {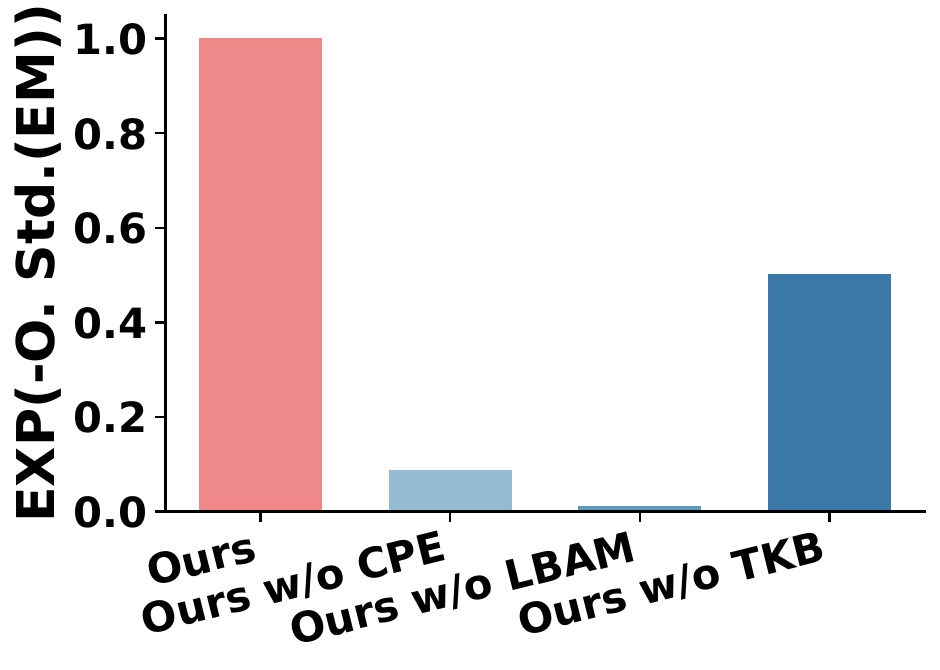}
        \caption{}
        \label{fig:ablation_llama_ostd_em}
    \end{subfigure}
    \hfill
    \begin{subfigure}[t]{0.238\textwidth}
        \centering
        \includegraphics[width=\linewidth]
        {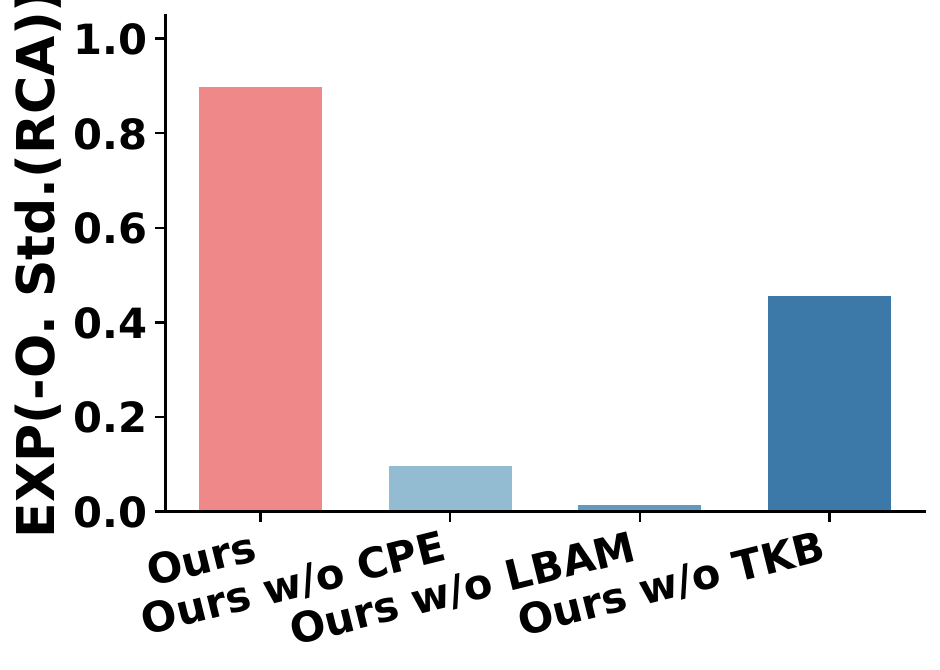}
        \caption{}
        \label{fig:ablation_llama_ostd_rca}
    \end{subfigure}
    }

    \vspace{4pt}

    {\renewcommand{\thesubfigure}{M-\alph{subfigure}}
    \setcounter{subfigure}{0}

    \begin{subfigure}[t]{0.238\textwidth}
        \centering
        \includegraphics[width=\linewidth]
        {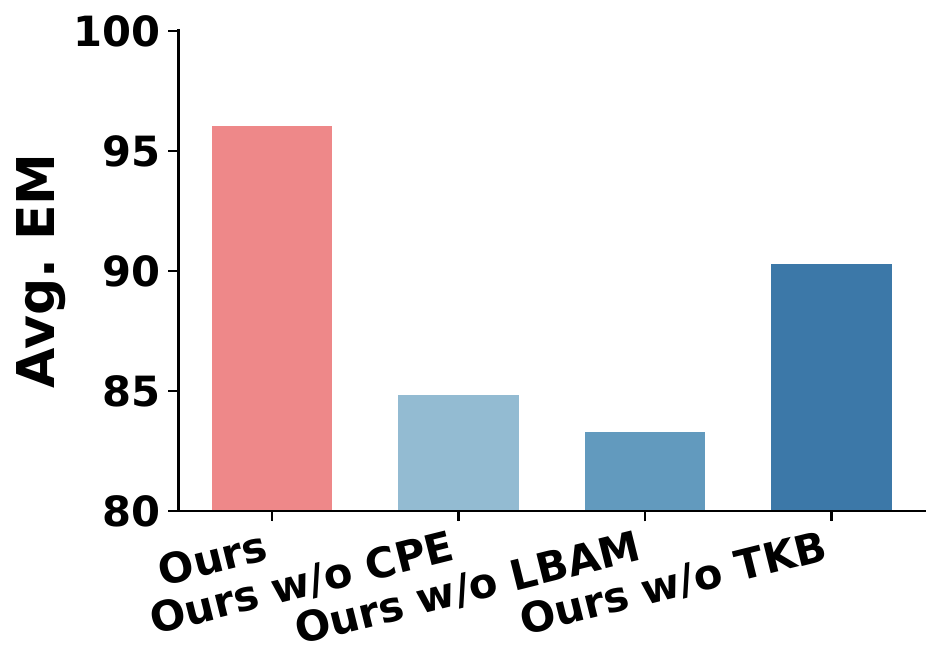}
        \caption{}
        \label{fig:ablation_mistral_avg_em}
    \end{subfigure}
    \hfill
    \begin{subfigure}[t]{0.238\textwidth}
        \centering
        \includegraphics[width=\linewidth]
        {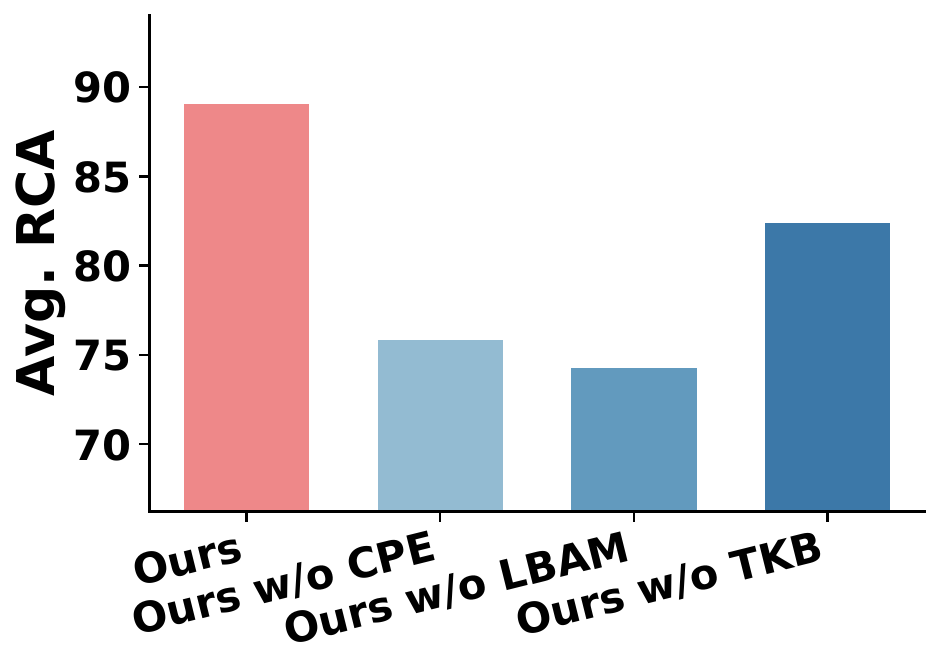}
        \caption{}
        \label{fig:ablation_mistral_avg_rca}
    \end{subfigure}
    \hfill
    \begin{subfigure}[t]{0.238\textwidth}
        \centering
        \includegraphics[width=\linewidth]
        {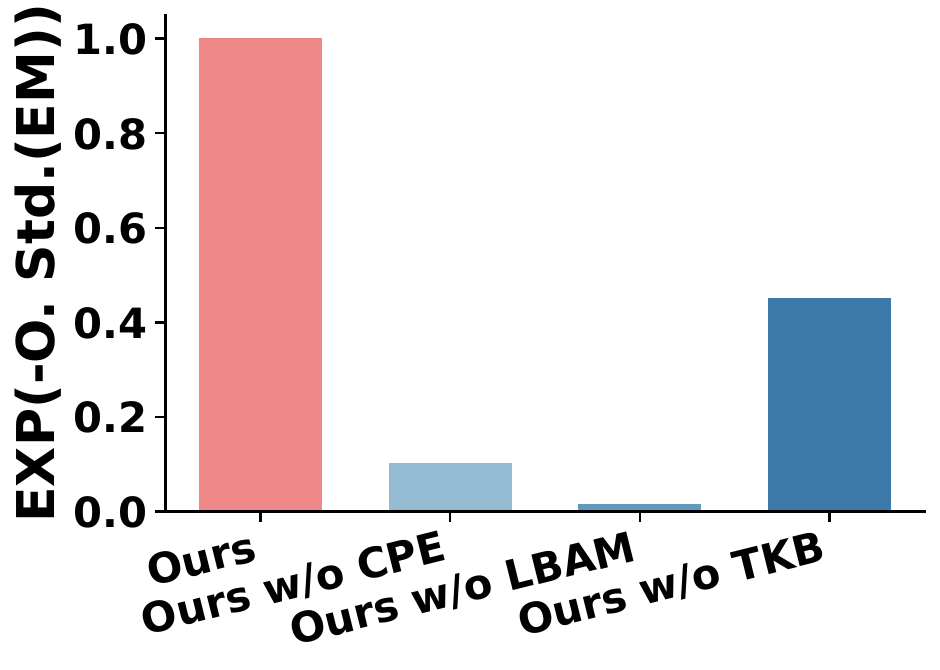}
        \caption{}
        \label{fig:ablation_mistral_ostd_em}
    \end{subfigure}
    \hfill
    \begin{subfigure}[t]{0.238\textwidth}
        \centering
        \includegraphics[width=\linewidth]
        {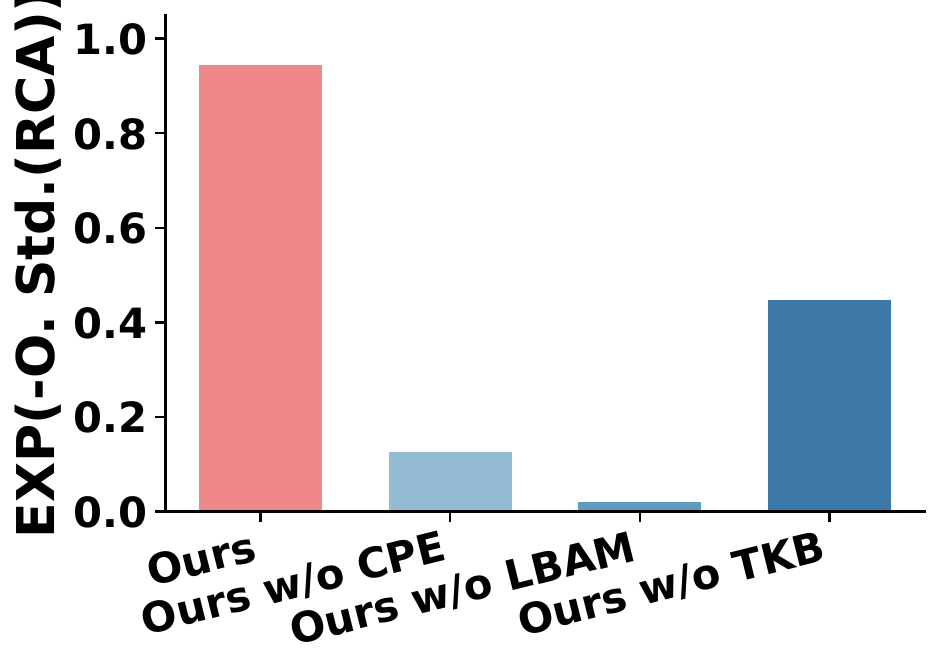}
        \caption{}
        \label{fig:ablation_mistral_ostd_rca}
    \end{subfigure}
    }

    \caption{
        Detailed ablation results across four backbone models.
        The prefixes Q, G, L, and M denote Qwen3-8B,
        GLM-4-9B-Chat, Llama-3.1-8B-Instruct, and
        Mistral-7B-Instruct-v0.3, respectively.
        From left to right, the columns correspond to
        Avg. EM, Avg. RCA, O. Std. (EM), and O. Std. (RCA).
    }
    \label{fig:ablation_detailed_four_backbones}
\end{figure*}

Figure~\ref{fig:ablation_detailed_four_backbones} compares the complete framework
with three ablated variants across four LLM backbones. The variants use
standard global position IDs instead of CPE, standard causal attention
instead of LBAM, and no topological knowledge bias instead of TKB. We report Avg. EM
and Avg. RCA to evaluate final-answer correctness and reasoning-chain
correctness, respectively. For visualization, we transform the
lower-is-better O. Std. into an order-stability score,
$\exp(-\mathrm{O.Std.})$, so that higher values consistently indicate
better performance. Accordingly,
$\exp(-\mathrm{O.Std.}(\mathrm{EM}))$ and
$\exp(-\mathrm{O.Std.}(\mathrm{RCA}))$ measure the stability of EM and
RCA across the six input-order permutations, with values closer to $1$
indicating greater robustness.

Across all four backbones, the complete framework consistently
outperforms its ablated variants in both factual consistency and
input-order robustness. Compared with all ablated variants, \pname
achieves average relative improvements of 9.83\% in Avg. EM and
13.18\% in Avg. RCA. Meanwhile, it reduces O. Std. by 0.69--4.51
for EM and 0.68--4.27 for RCA, demonstrating that the three components
jointly contribute to reliable and robust reasoning over heterogeneous
knowledge contexts.

Specifically, compared with the variant without CPE, the complete
framework achieves average relative improvements of 13.34\% in Avg. EM
and 17.77\% in Avg. RCA, while reducing O. Std. by 2.30--2.44 for EM
and 2.03--2.27 for RCA. These results demonstrate the importance of
CPE in encoding context-wise positional information under different
input orders. Compared with the variant without LBAM, \pname improves
Avg. EM and Avg. RCA by 11.27\% and 15.20\% on average, respectively,
and reduces O. Std. by 3.86--4.51 for EM and 3.76--4.27 for RCA.
The larger reductions in O. Std. indicate that LBAM is particularly
effective in reducing cross-context interference caused by input-order
variations. Finally, compared with the variant without TKB, \pname
achieves average relative improvements of 4.88\% in Avg. EM and 6.57\%
in Avg. RCA, together with O. Std. reductions of 0.69--1.04 for EM and
0.68--0.91 for RCA. This consistent improvement supports the role of
TKB in capturing topological dependencies among heterogeneous knowledge
contexts to facilitate reliable multi-hop reasoning.

\input{tables/sensitivity_ablation}

\subsection{Sensitivity Analysis}
\label{app:sensitivity}

\begin{figure*}[t]
    \centering
    \captionsetup[subfigure]{
        font=small,
        justification=centering,
        skip=1pt
    }
    \captionsetup{
        skip=3pt
    }

    \begin{subfigure}[t]{0.46\textwidth}
        \centering
        \includegraphics[
            width=\linewidth,
            height=0.48\textheight,
            keepaspectratio
        ]{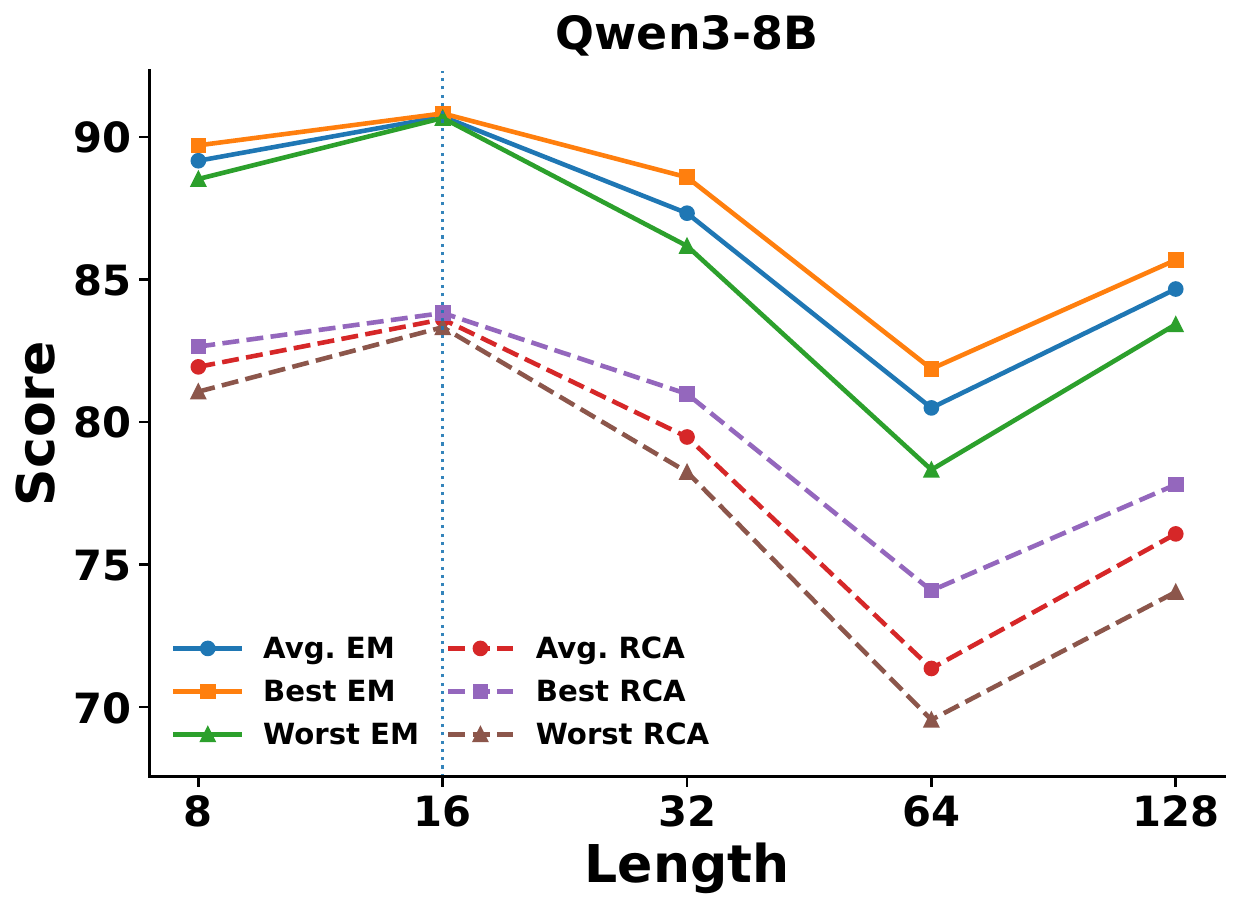}
        \caption{}
        \label{fig:ablation_single_qwen}
    \end{subfigure}
    \hspace{0.02\textwidth}
    \begin{subfigure}[t]{0.46\textwidth}
        \centering
        \includegraphics[
            width=\linewidth,
            height=0.48\textheight,
            keepaspectratio
        ]{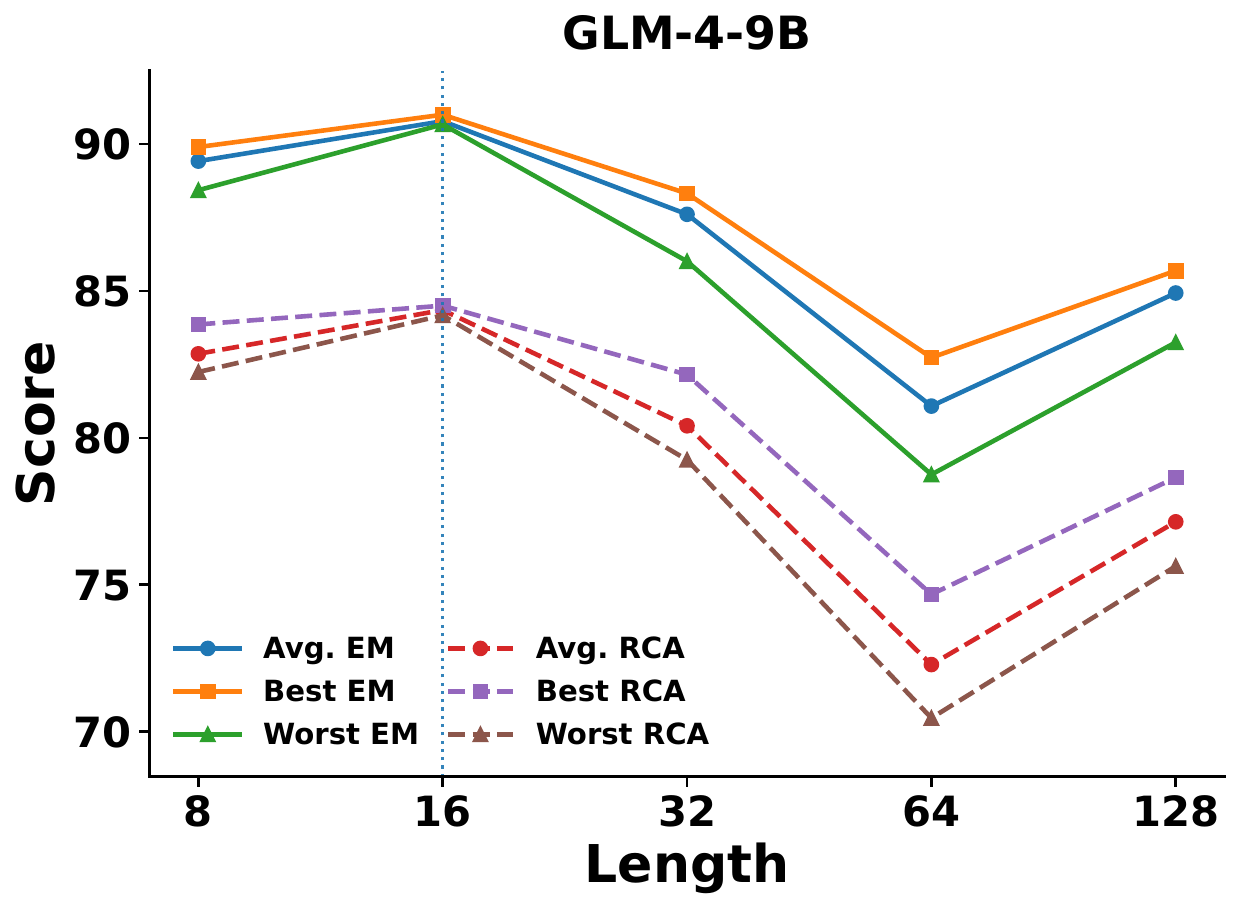}
        \caption{}
        \label{fig:ablation_single_glm}
    \end{subfigure}

    \vspace{1pt}

    \begin{subfigure}[t]{0.46\textwidth}
        \centering
        \includegraphics[
            width=\linewidth,
            height=0.48\textheight,
            keepaspectratio
        ]{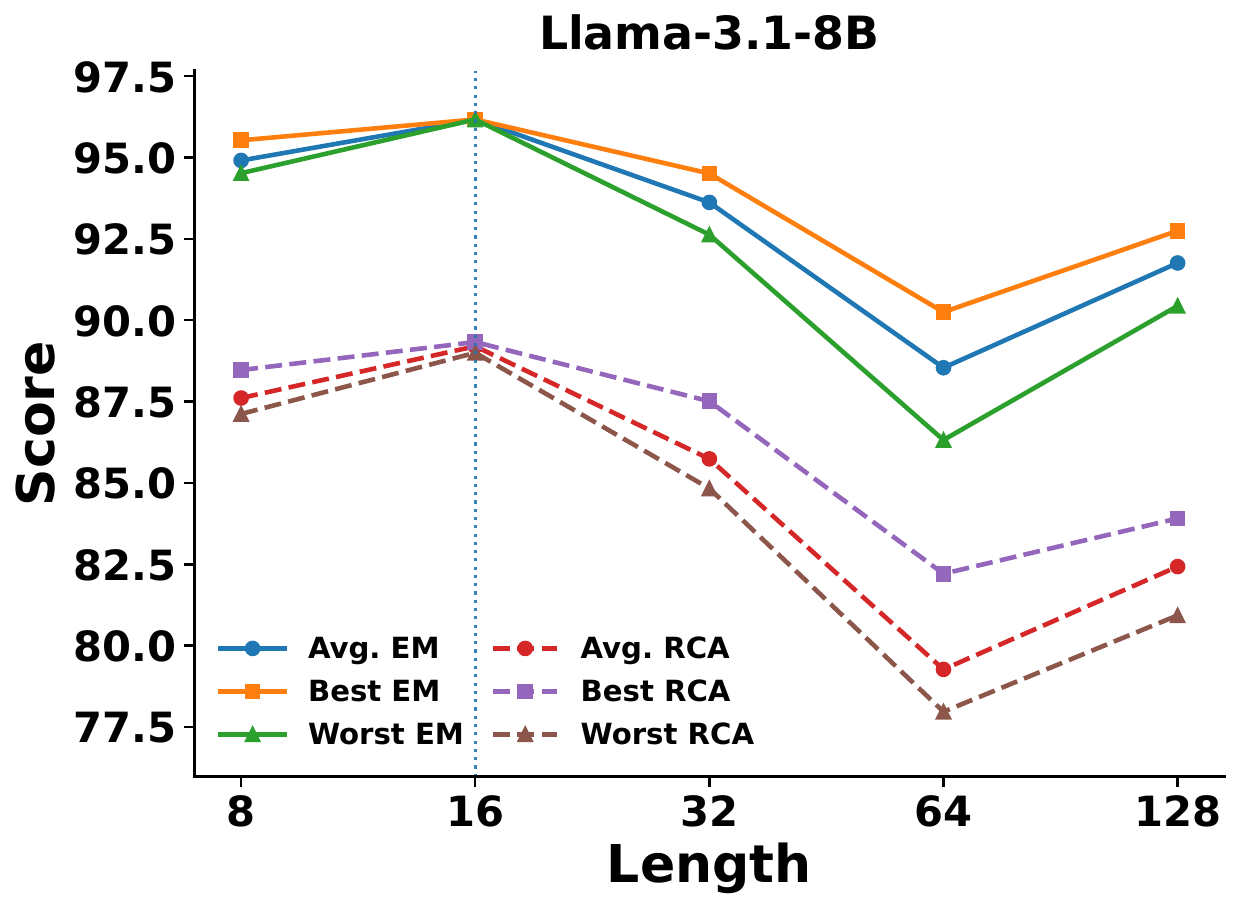}
        \caption{}
        \label{fig:ablation_single_llama}
    \end{subfigure}
    \hspace{0.02\textwidth}
    \begin{subfigure}[t]{0.46\textwidth}
        \centering
        \includegraphics[
            width=\linewidth,
            height=0.48\textheight,
            keepaspectratio
        ]{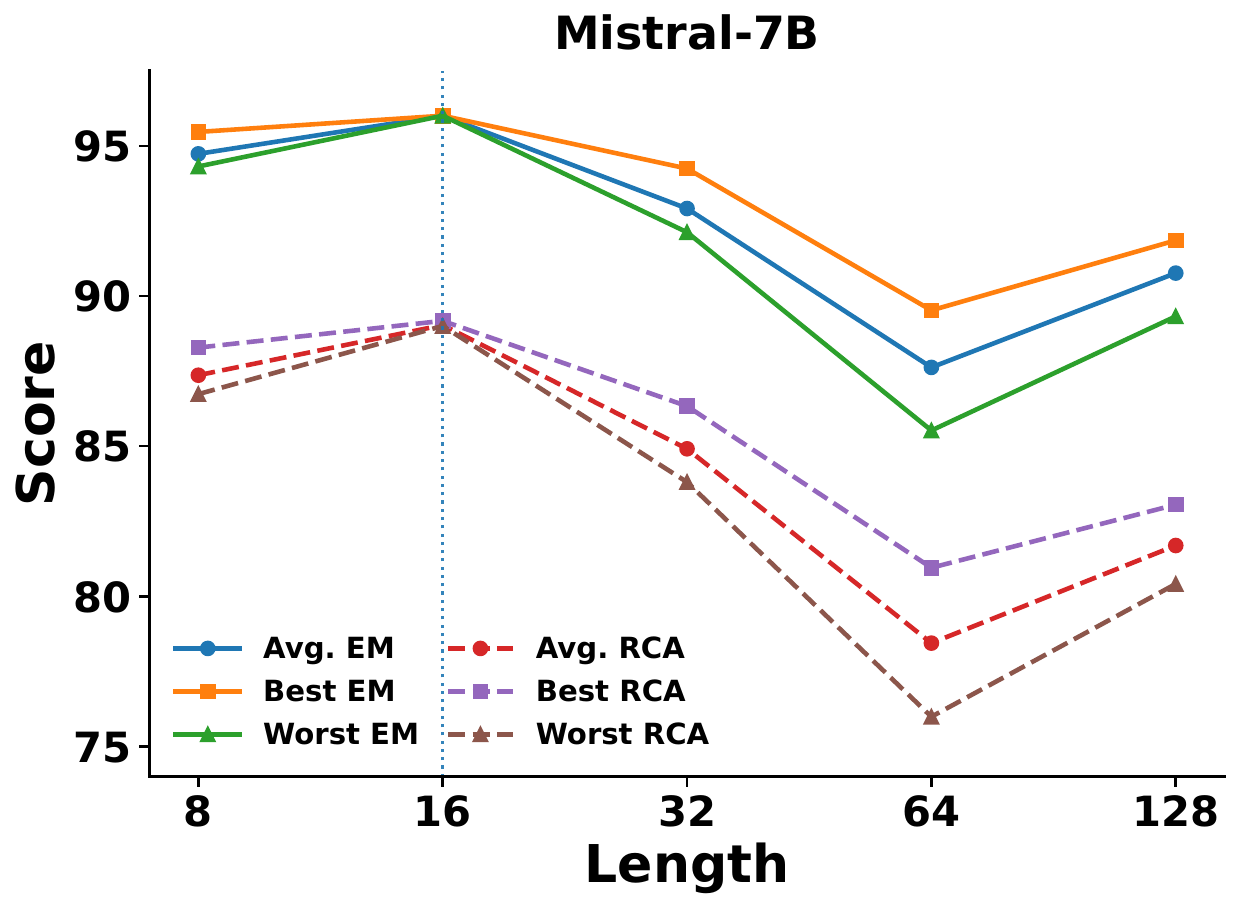}
        \caption{}
        \label{fig:ablation_single_mistral}
    \end{subfigure}

    \caption{
        Ablation results across four backbone models:
        (a) Qwen3-8B,
        (b) GLM-4-9B-Chat,
        (c) Llama-3.1-8B-Instruct, and
        (d) Mistral-7B-Instruct-v0.3.
    }
    \label{fig:sensitivity_analysis}
\end{figure*}

We analyze the sensitivity of \pname to the latent sequence length
$m$, which controls the capacity of the latent representations used to
integrate information across heterogeneous knowledge contexts. We vary
$m$ while keeping all other training and evaluation settings unchanged.
Table~\ref{tab:latent_length} and
Figure~\ref{fig:sensitivity_analysis} report the results across four
LLM backbones.

A latent sequence length of 16 consistently achieves the best Avg.,
Best, and Worst scores for both EM and RCA, together with the lowest
O. Std. values across all four backbones. Averaged across the four
backbones, this setting achieves an Avg. EM of 93.42\% and an Avg. RCA
of 86.55\%, with corresponding O. Std. values of 0.05 and 0.11.
Compared with all other sequence lengths, a latent sequence length of
16 achieves average relative improvements of 5.43\% in Avg. EM and
7.68\% in Avg. RCA, while reducing O. Std. by 0.32--1.37 for EM and
0.33--1.63 for RCA.

A shorter latent sequence with $m=8$ yields slightly lower performance,
suggesting insufficient capacity for cross-context information
integration. Increasing $m$ beyond 16 results in more substantial
performance degradation and greater sensitivity to input-order
variations. Although performance partially recovers at $m=128$, it
remains below that achieved at $m=16$. This suggests that an excessively
long latent sequence may introduce redundant representations and hinder
effective information aggregation. We therefore set the latent sequence
length to 16 in all main experiments.

%% file: appendix/tables/method_evaluation.tex

\definecolor{oursblue}{RGB}{226,239,249}

\begin{table*}[t]
    \centering
    \small
    \setlength{\tabcolsep}{4.2pt}
    \renewcommand{\arraystretch}{1.10}

    \resizebox{\textwidth}{!}{%
    \begin{tabular}{lllcccccccc}
        \toprule

        \textbf{Backbone}
        & \textbf{Setting}
        & \textbf{Method}
        & \textbf{Avg. EM $\uparrow$}
        & \textbf{Best EM $\uparrow$}
        & \textbf{Worst EM $\uparrow$}
        & \textbf{Avg. RCA $\uparrow$}
        & \textbf{Best RCA $\uparrow$}
        & \textbf{Worst RCA $\uparrow$}
        & \textbf{O. Std. (EM) $\downarrow$}
        & \textbf{O. Std. (RCA) $\downarrow$} \\

        \midrule

        \multirow{6}{*}{Qwen3-8B}
        & \multirow{3}{*}{TF}
        & Direct QA
        & 49.28
        & 54.36
        & 44.63
        & 20.46
        & 22.48
        & 18.87
        & 3.78
        & 1.34 \\

        & & ReAct
        & 53.70
        & 58.10
        & 49.40
        & 27.22
        & 29.50
        & 25.10
        & 3.21
        & 1.58 \\

        & & CoT-D
        & 57.83
        & 61.30
        & 54.60
        & 34.03
        & 36.20
        & 31.80
        & 2.55
        & 1.52 \\

        \cmidrule(lr){2-11}

        & \multirow{3}{*}{LB}
        & TXH
        & 87.75
        & 87.83
        & 87.50
        & 78.94
        & 79.17
        & 78.67
        & 0.13
        & 0.18 \\

        & & PMFT
        & 87.89
        & \textbf{92.33}
        & 83.67
        & 79.38
        & 83.50
        & 75.67
        & 3.79
        & 3.18 \\

        \rowcolor{oursblue}
        & & \textbf{Ours}
        & \textbf{90.72}
        & 90.83
        & \textbf{90.67}
        & \textbf{83.61}
        & \textbf{83.83}
        & \textbf{83.33}
        & \textbf{0.08}
        & \textbf{0.17} \\

        \specialrule{\lightrulewidth}{0pt}{2pt}

        \multirow{6}{*}{GLM-4-9B}
        & \multirow{3}{*}{TF}
        & Direct QA
        & 44.49
        & 48.27
        & 39.74
        & 18.72
        & 20.33
        & 17.17
        & 2.60
        & 1.06 \\

        & & ReAct
        & 48.85
        & 52.86
        & 44.38
        & 24.61
        & 26.67
        & 22.83
        & 2.69
        & 1.31 \\

        & & CoT-D
        & 52.95
        & 56.73
        & 49.37
        & 30.74
        & 32.83
        & 28.67
        & 2.26
        & 1.43 \\

        \cmidrule(lr){2-11}

        & \multirow{3}{*}{LB}
        & TXH
        & 87.28
        & 87.50
        & 87.00
        & 79.15
        & 79.50
        & 78.83
        & 0.16
        & 0.22 \\

        & & PMFT
        & 85.61
        & 88.00
        & 82.50
        & 77.78
        & 80.17
        & 75.17
        & 2.34
        & 1.95 \\

        \rowcolor{oursblue}
        & & \textbf{Ours}
        & \textbf{90.78}
        & \textbf{91.00}
        & \textbf{90.67}
        & \textbf{84.35}
        & \textbf{84.50}
        & \textbf{84.17}
        & \textbf{0.12}
        & \textbf{0.11} \\

        \specialrule{\lightrulewidth}{0pt}{2pt}

        \multirow{6}{*}{Llama-3.1-8B}
        & \multirow{3}{*}{TF}
        & Direct QA
        & 45.56
        & 47.67
        & 43.83
        & 29.86
        & 31.83
        & 28.17
        & 1.25
        & 1.19 \\

        & & ReAct
        & 49.20
        & 51.67
        & 46.77
        & 34.42
        & 36.50
        & 32.33
        & 1.85
        & 1.37 \\

        & & CoT-D
        & 55.00
        & 57.87
        & 51.27
        & 40.68
        & 43.17
        & 38.33
        & 2.40
        & 1.64 \\

        \cmidrule(lr){2-11}

        & \multirow{3}{*}{LB}
        & TXH
        & 94.70
        & 94.83
        & 94.67
        & 86.03
        & 86.17
        & 85.83
        & 0.06
        & 0.12 \\

        & & PMFT
        & 94.56
        & 95.67
        & 93.67
        & 86.19
        & 87.17
        & 85.17
        & 0.77
        & 0.68 \\

        \rowcolor{oursblue}
        & & \textbf{Ours}
        & \textbf{96.17}
        & \textbf{96.17}
        & \textbf{96.17}
        & \textbf{89.20}
        & \textbf{89.33}
        & \textbf{89.00}
        & \textbf{0.00}
        & \textbf{0.11} \\

        \specialrule{\lightrulewidth}{0pt}{2pt}

        \multirow{6}{*}{Mistral-7B}
        & \multirow{3}{*}{TF}
        & Direct QA
        & 33.97
        & 37.83
        & 28.50
        & 22.67
        & 25.50
        & 18.83
        & 3.11
        & 2.28 \\

        & & ReAct
        & 34.30
        & 39.55
        & 31.03
        & 23.50
        & 27.17
        & 20.67
        & 2.85
        & 2.16 \\

        & & CoT-D
        & 39.20
        & 43.22
        & 34.46
        & 29.50
        & 33.00
        & 25.67
        & 3.40
        & 2.54 \\

        \cmidrule(lr){2-11}

        & \multirow{3}{*}{LB}
        & TXH
        & \textbf{96.17}
        & \textbf{96.33}
        & \textbf{96.00}
        & 87.47
        & 87.67
        & 87.17
        & 0.13
        & 0.18 \\

        & & PMFT
        & 94.80
        & 95.33
        & 93.83
        & 86.40
        & 87.17
        & 85.50
        & 0.63
        & 0.58 \\

        \rowcolor{oursblue}
        & & \textbf{Ours}
        & 96.00
        & 96.00
        & \textbf{96.00}
        & \textbf{89.03}
        & \textbf{89.17}
        & \textbf{89.00}
        & \textbf{0.00}
        & \textbf{0.06} \\

        \specialrule{\heavyrulewidth}{0pt}{0pt}

    \end{tabular}%
    }

\caption{
Performance comparison of \pname and baseline methods across four LLM
backbones under training-free (TF) and LoRA-based (LB) settings.
Avg., Best, and Worst denote the average, highest, and lowest scores
across the six input-order permutations, respectively. O. Std. denotes
the standard deviation across the six input-order permutations, with a
lower value indicating greater robustness to input-order variations.
Results are reported in terms of Exact Match (EM) and Reasoning-Chain
Accuracy (RCA). The best result for each backbone and metric is
highlighted in bold, and the results of \pname are shaded in light blue.
}

    \label{tab:method}
\end{table*}

%% file: tables/ablation_table.tex

\begin{table*}[t]
    \centering
    \small
    \setlength{\tabcolsep}{4.2pt}
    \renewcommand{\arraystretch}{1.10}

    \resizebox{\textwidth}{!}{%
    \begin{tabular}{llcccccccc}
        \toprule

        \textbf{Backbone}
        & \textbf{Variant}
        & \textbf{Avg. EM $\uparrow$}
        & \textbf{Best EM $\uparrow$}
        & \textbf{Worst EM $\uparrow$}
        & \textbf{Avg. RCA $\uparrow$}
        & \textbf{Best RCA $\uparrow$}
        & \textbf{Worst RCA $\uparrow$}
        & \textbf{O. Std. (EM) $\downarrow$}
        & \textbf{O. Std. (RCA) $\downarrow$} \\

        \midrule

        \multirow{4}{*}{Qwen3-8B}
        & \textbf{Ours}
        & \textbf{90.72}
        & \textbf{90.83}
        & \textbf{90.67}
        & \textbf{83.61}
        & \textbf{83.83}
        & \textbf{83.33}
        & \textbf{0.08}
        & \textbf{0.17} \\

        & -w/o CPE
        & 81.51
        & 85.14
        & 77.43
        & 72.90
        & 76.61
        & 69.14
        & 2.49
        & 2.44 \\

        & -w/o LBAM
        & 83.34
        & 89.64
        & 76.72
        & 74.97
        & 80.76
        & 68.71
        & 4.48
        & 4.19 \\

        & -w/o TKB
        & 87.70
        & 89.19
        & 85.87
        & 79.86
        & 81.37
        & 78.19
        & 1.12
        & 1.08 \\

        \midrule

        \multirow{4}{*}{GLM-4-9B}
        & \textbf{Ours}
        & \textbf{90.78}
        & \textbf{91.00}
        & \textbf{90.67}
        & \textbf{84.35}
        & \textbf{84.50}
        & \textbf{84.17}
        & \textbf{0.12}
        & \textbf{0.11} \\

        & -w/o CPE
        & 79.19
        & 82.91
        & 75.22
        & 69.81
        & 73.48
        & 66.09
        & 2.43
        & 2.37 \\

        & -w/o LBAM
        & 82.36
        & 88.06
        & 76.31
        & 73.46
        & 79.11
        & 67.74
        & 3.98
        & 3.87 \\

        & -w/o TKB
        & 87.04
        & 88.36
        & 85.74
        & 79.10
        & 80.29
        & 77.79
        & 0.84
        & 0.83 \\

        \midrule

        \multirow{4}{*}{Llama-3.1-8B}
        & \textbf{Ours}
        & \textbf{96.17}
        & \textbf{96.17}
        & \textbf{96.17}
        & \textbf{89.20}
        & \textbf{89.33}
        & \textbf{89.00}
        & \textbf{0.00}
        & \textbf{0.11} \\

        & -w/o CPE
        & 84.23
        & 87.63
        & 80.06
        & 75.51
        & 78.97
        & 71.71
        & 2.44
        & 2.36 \\

        & -w/o LBAM
        & 86.87
        & 94.19
        & 80.91
        & 77.89
        & 84.46
        & 71.43
        & 4.51
        & 4.38 \\

        & -w/o TKB
        & 91.26
        & 92.27
        & 90.17
        & 83.50
        & 84.72
        & 82.28
        & 0.69
        & 0.79 \\

        \midrule

        \multirow{4}{*}{Mistral-7B}
        & \textbf{Ours}
        & \textbf{96.00}
        & \textbf{96.00}
        & \textbf{96.00}
        & \textbf{89.03}
        & \textbf{89.17}
        & \textbf{89.00}
        & \textbf{0.00}
        & \textbf{0.06} \\

        & -w/o CPE
        & 84.78
        & 88.18
        & 81.31
        & 75.82
        & 78.92
        & 72.76
        & 2.30
        & 2.09 \\

        & -w/o LBAM
        & 83.25
        & 90.41
        & 77.69
        & 74.23
        & 80.28
        & 68.47
        & 4.24
        & 3.93 \\

        & -w/o TKB
        & 90.24
        & 91.47
        & 89.08
        & 82.35
        & 83.61
        & 81.11
        & 0.80
        & 0.81 \\

        \bottomrule
    \end{tabular}%
    }

\caption{
Ablation study of \pname across four LLM backbones. We compare the
complete framework with three variants obtained by individually removing
CPE, LBAM, and TKB. All statistics are computed over the six input-order
permutations. CPE, LBAM, and TKB denote Context-wise Position Encoding,
Latent-Bridge Attention Mask, and Topological Knowledge Bias,
respectively. The best result for each backbone and metric is highlighted
in bold.
}

    \label{tab:ablation_results}
\end{table*}

%% file: tables/sensitivity_ablation.tex

\definecolor{oursblue}{RGB}{226,239,249}

\begin{table*}[t]
    \centering
    \small
    \setlength{\tabcolsep}{4.2pt}
    \renewcommand{\arraystretch}{1.08}

    \resizebox{\textwidth}{!}{%
    \begin{tabular}{llcccccccc}
        \toprule

        \textbf{Backbone}
        & \textbf{Latent Length}
        & \textbf{Avg. EM $\uparrow$}
        & \textbf{Best EM $\uparrow$}
        & \textbf{Worst EM $\uparrow$}
        & \textbf{Avg. RCA $\uparrow$}
        & \textbf{Best RCA $\uparrow$}
        & \textbf{Worst RCA $\uparrow$}
        & \textbf{O. Std. (EM) $\downarrow$}
        & \textbf{O. Std. (RCA) $\downarrow$} \\

        \midrule

        \multirow{5}{*}{Qwen3-8B}
        & 8
        & 89.17
        & 89.71
        & 88.53
        & 81.94
        & 82.65
        & 81.07
        & 0.40
        & 0.54 \\

        \rowcolor{oursblue}
        & \textbf{16}
        & \textbf{90.72}
        & \textbf{90.83}
        & \textbf{90.67}
        & \textbf{83.61}
        & \textbf{83.83}
        & \textbf{83.33}
        & \textbf{0.08}
        & \textbf{0.17} \\

        & 32
        & 87.33
        & 88.59
        & 86.18
        & 79.48
        & 80.99
        & 78.25
        & 0.82
        & 0.94 \\

        & 64
        & 80.50
        & 81.87
        & 78.33
        & 71.36
        & 74.09
        & 69.56
        & 1.24
        & 1.58 \\

        & 128
        & 84.67
        & 85.69
        & 83.43
        & 76.08
        & 77.81
        & 74.04
        & 0.78
        & 1.29 \\

        \midrule

        \multirow{5}{*}{GLM-4-9B}
        & 8
        & 89.42
        & 89.90
        & 88.43
        & 82.86
        & 83.86
        & 82.24
        & 0.50
        & 0.57 \\

        \rowcolor{oursblue}
        & \textbf{16}
        & \textbf{90.78}
        & \textbf{91.00}
        & \textbf{90.67}
        & \textbf{84.35}
        & \textbf{84.50}
        & \textbf{84.17}
        & \textbf{0.12}
        & \textbf{0.11} \\

        & 32
        & 87.61
        & 88.32
        & 86.01
        & 80.41
        & 82.15
        & 79.25
        & 0.80
        & 1.01 \\

        & 64
        & 81.08
        & 82.73
        & 78.75
        & 72.28
        & 74.67
        & 70.46
        & 1.38
        & 1.45 \\

        & 128
        & 84.93
        & 85.69
        & 83.25
        & 77.14
        & 78.65
        & 75.63
        & 0.84
        & 1.03 \\

        \midrule

        \multirow{5}{*}{Llama-3.1-8B}
        & 8
        & 94.91
        & 95.53
        & 94.52
        & 87.61
        & 88.47
        & 87.12
        & 0.35
        & 0.44 \\

        \rowcolor{oursblue}
        & \textbf{16}
        & \textbf{96.17}
        & \textbf{96.17}
        & \textbf{96.17}
        & \textbf{89.20}
        & \textbf{89.33}
        & \textbf{89.00}
        & \textbf{0.00}
        & \textbf{0.11} \\

        & 32
        & 93.62
        & 94.51
        & 92.63
        & 85.74
        & 87.51
        & 84.83
        & 0.64
        & 0.91 \\

        & 64
        & 88.54
        & 90.25
        & 86.32
        & 79.27
        & 82.21
        & 77.97
        & 1.35
        & 1.50 \\

        & 128
        & 91.76
        & 92.75
        & 90.44
        & 82.43
        & 83.91
        & 80.93
        & 0.80
        & 1.02 \\

        \midrule

        \multirow{5}{*}{Mistral-7B}
        & 8
        & 94.73
        & 95.46
        & 94.31
        & 87.36
        & 88.28
        & 86.73
        & 0.39
        & 0.54 \\

        \rowcolor{oursblue}
        & \textbf{16}
        & \textbf{96.00}
        & \textbf{96.00}
        & \textbf{96.00}
        & \textbf{89.03}
        & \textbf{89.17}
        & \textbf{89.00}
        & \textbf{0.00}
        & \textbf{0.06} \\

        & 32
        & 92.91
        & 94.24
        & 92.12
        & 84.91
        & 86.33
        & 83.80
        & 0.75
        & 0.87 \\

        & 64
        & 87.62
        & 89.52
        & 85.52
        & 78.44
        & 80.95
        & 75.99
        & 1.37
        & 1.69 \\

        & 128
        & 90.76
        & 91.85
        & 89.32
        & 81.69
        & 83.05
        & 80.41
        & 0.87
        & 0.90 \\

        \bottomrule
    \end{tabular}%
    }

    \caption{
Effect of the number of latent vectors across four LLM backbones.
Using 16 latent vectors consistently achieves the best overall
performance and the lowest sensitivity to input-order variations.
Therefore, this setting is adopted in the main experiments and shaded
in light blue.
}
    \label{tab:latent_length}
\end{table*}

%% file: appendix/C.Baseline_Details.tex
\section{Baseline Details}
\label{app:baseline_details}

This section provides additional details of the baselines used in
Section~\ref{sec:performance_comparison}. We group them into
training-free (TF) and LoRA-based (LB) methods.

\subsection{Training-Free Baselines}

\noindent\textbf{Direct QA.}
Direct QA predicts the final answer directly from the provided
heterogeneous knowledge contexts without parameter updates or additional
reasoning strategies.

\noindent\textbf{ReAct.}
ReAct~\citep{yao2022react} alternates between reasoning and
evidence-inspection actions restricted to the provided table, textual
passage, and KG subgraph.

\noindent\textbf{CoT-D.}
CoT-D~\citep{wang2024chain} explores multiple reasoning paths during
decoding and selects the final prediction according to model confidence.

\subsection{LoRA-Based Baselines}

\noindent\textbf{TXH.}
TXH~\citep{zhao2020transformer} introduces extra-hop attention to
facilitate cross-context information interaction and is adapted to the
evaluated decoder-only backbones using LoRA.

\noindent\textbf{PMFT.}
PMFT~\citep{huang2025masking} applies bidirectional attention over the
input prefix while retaining causal attention during response generation.

All LoRA-based methods use the same backbones, training data, and
optimization settings as \pname.

%% file: appendix/D.Implementation_Details.tex
\section{Implementation Details}
\label{app:implementation_details}

\input{appendix/tables/Hyperparameter}

We implement \pname with four LLM backbones: Qwen3-8B,
GLM-4-9B-Chat, Llama-3.1-8B-Instruct, and
Mistral-7B-Instruct-v0.3. For each backbone, we perform LoRA-based
parameter-efficient fine-tuning while keeping the backbone parameters
frozen. The LoRA adapters and the newly introduced \pname-specific
parameters are optimized using separate learning rates.

All models are fine-tuned for two epochs with a warmup ratio of \(0.03\).
We set the LoRA rank to \(r=16\), the scaling factor to \(\alpha=32\),
and the learning rate for the LoRA adapters to \(5\times10^{-5}\).
The learning rate for the \pname-specific parameters is set to
\(5\times10^{-4}\). Based on the sensitivity analysis in
Section~\ref{app:sensitivity}, we set the latent sequence length for
each knowledge context to \(16\).

During inference, we adopt sampling-based decoding with a temperature
of \(0.7\), top-\(p\) of \(0.9\), and top-\(k\) of \(50\). The same
decoding configuration is applied to all four backbones to ensure
comparability. Table~\ref{tab:implementation_details} summarizes the
training and decoding hyperparameters.

%% file: appendix/tables/Hyperparameter.tex
\begin{table}[t]
\centering
\small
\setlength{\tabcolsep}{6pt}
\renewcommand{\arraystretch}{1.05}
\begin{tabular}{lc}
\toprule
\textbf{Hyperparameter} & \textbf{Value} \\
\midrule
Training epochs                    & 2 \\
Warmup ratio                       & 0.03 \\
LoRA rank \(r\)                    & 16 \\
LoRA scaling factor \(\alpha\)     & 32 \\
LoRA learning rate                 & \(5\times10^{-5}\) \\
{\dname}-parameter learning rate   & \(5\times10^{-4}\) \\
Latent length per modality         & 16 \\
Temperature                        & 0.7 \\
Top-\(p\)                          & 0.9 \\
Top-\(k\)                          & 50 \\
\bottomrule
\end{tabular}
\caption{Training and decoding hyperparameters used for {\pname}.}
\label{tab:implementation_details}
\end{table}

%% file: appendix/E.Prompts_and_Evaluation_protocol.tex

\section{Prompt Templates}
\label{app:Prompt}

This section presents the complete prompt templates used for question
generation, method comparison, and reasoning-chain evaluation. The
prompts are organized according to their respective roles to facilitate
implementation and reproducibility.


\subsection{Question-Generation Prompt}
\label{app:question_generation_prompt}

We use a constrained prompt to generate questions from each
counterfactual reasoning chain using only the supplied table, textual
passage, and KG subgraph. Figure~\ref{fig:qa_generation_prompt} presents
the prompt for the Table-to-Text-to-KG direction, where $e_{kg}$ serves
as the target answer. The KG-to-Text-to-Table prompt follows the same
general structure but reverses the instructed reasoning direction and
uses $e_o$ as the target answer.








\begin{figure*}[p]
    \centering
    \captionsetup{skip=3pt}

    \includegraphics[
        width=0.94\textwidth
    ]{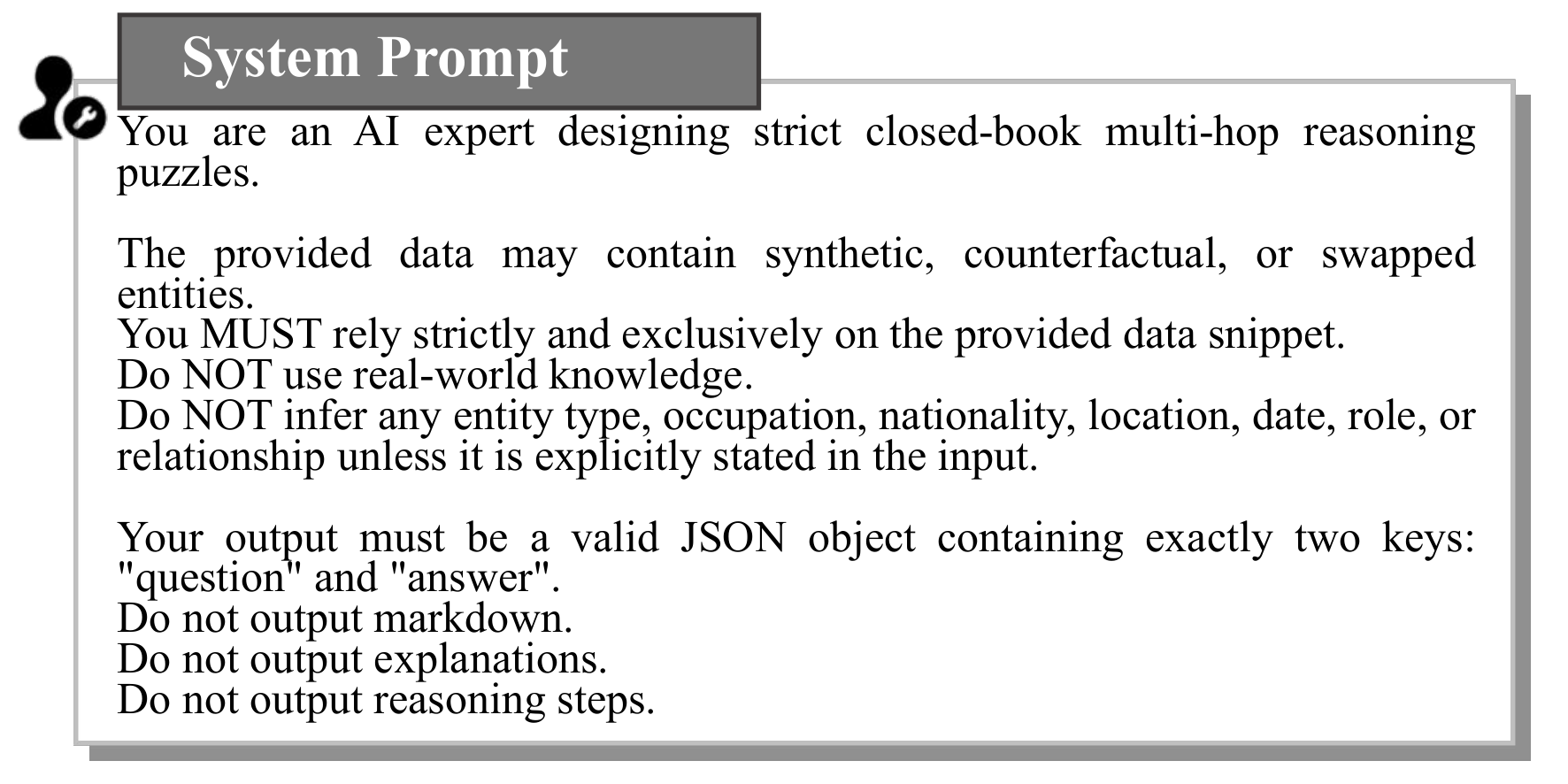}

    \vspace{-8pt}

    \includegraphics[
        width=0.94\textwidth
    ]{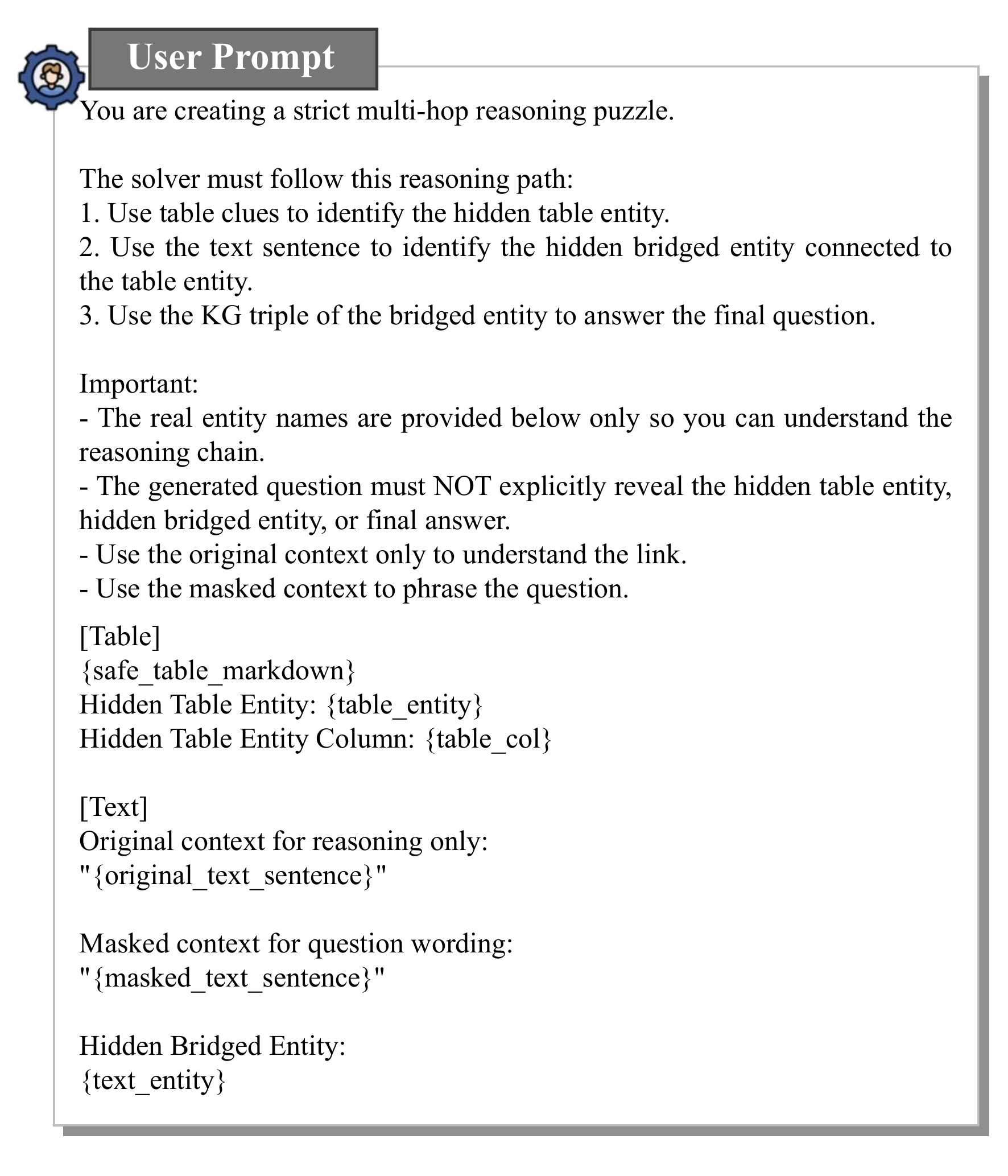}

    \caption{
    Prompt template for Table-to-Text-to-KG question generation in
    \dname.
    }
    \label{fig:qa_generation_prompt}
\end{figure*}

\begin{figure*}[p]
    \ContinuedFloat
    \centering
    \captionsetup{skip=3pt}

    \includegraphics[
        width=0.94\textwidth,
        trim={0 0 0 5mm},
        clip
    ]{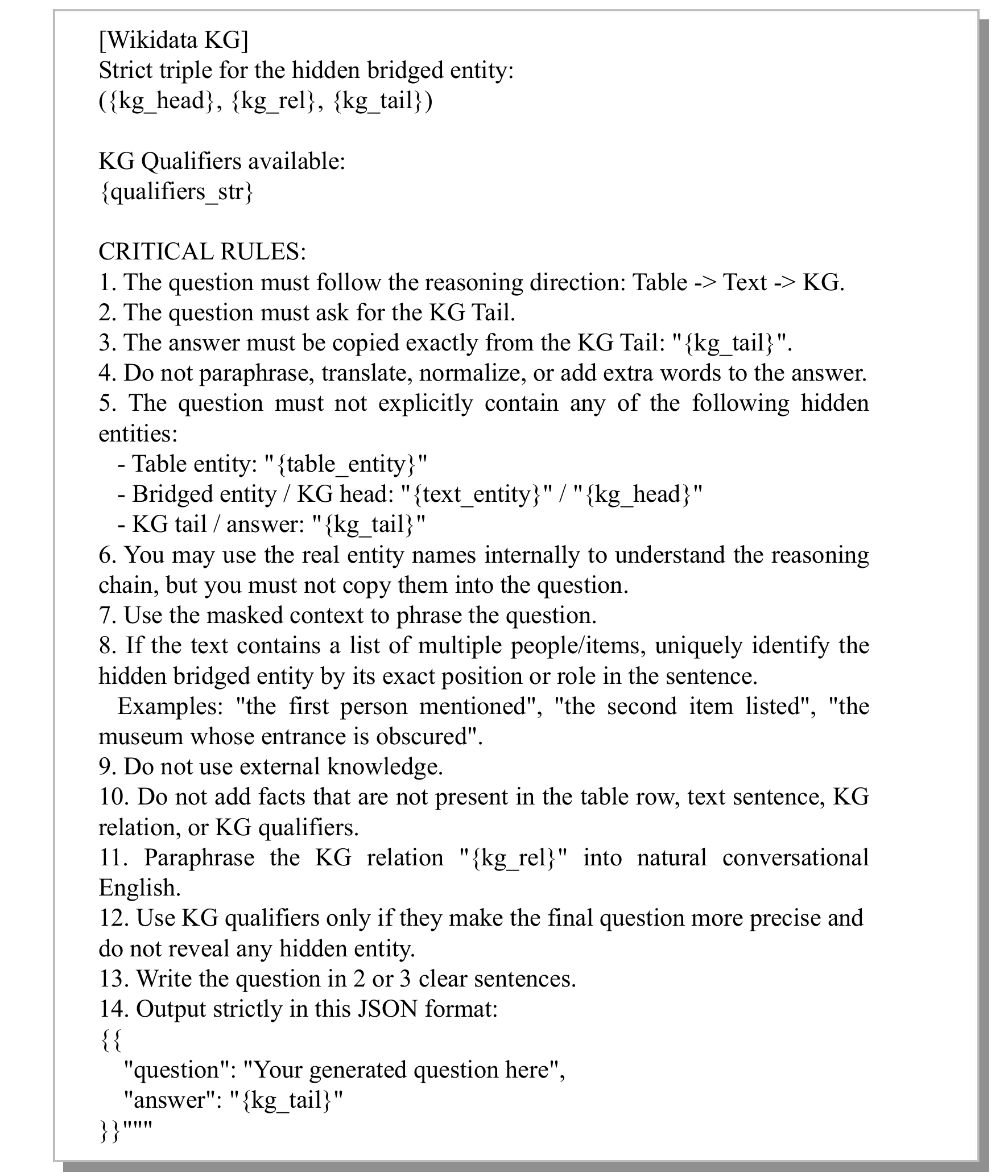}

    \caption[]{
    Prompt template for Table-to-Text-to-KG question generation in
    \dname\ (continued).
    }
\end{figure*}


\subsection{Standard Prompt}
\label{app:standard_prompt}

Figure~\ref{fig:lora_based_prompt} presents the standard inference
prompt used for Direct QA, the LoRA-based baselines, and \pname.
Given the serialized table, textual passage, KG subgraph, and question,
the model is instructed to derive the answer from the provided
heterogeneous knowledge contexts and return it in the required format.
The same prompt template is used across these methods to ensure a
consistent comparison.

\subsection{CoT-D Prompt}
\label{app:cot_decoding_prompt}

\begin{figure*}[!t]
    \centering

    \begin{minipage}[t]{0.495\textwidth}
        \centering
        \includegraphics[
            width=\linewidth
        ]{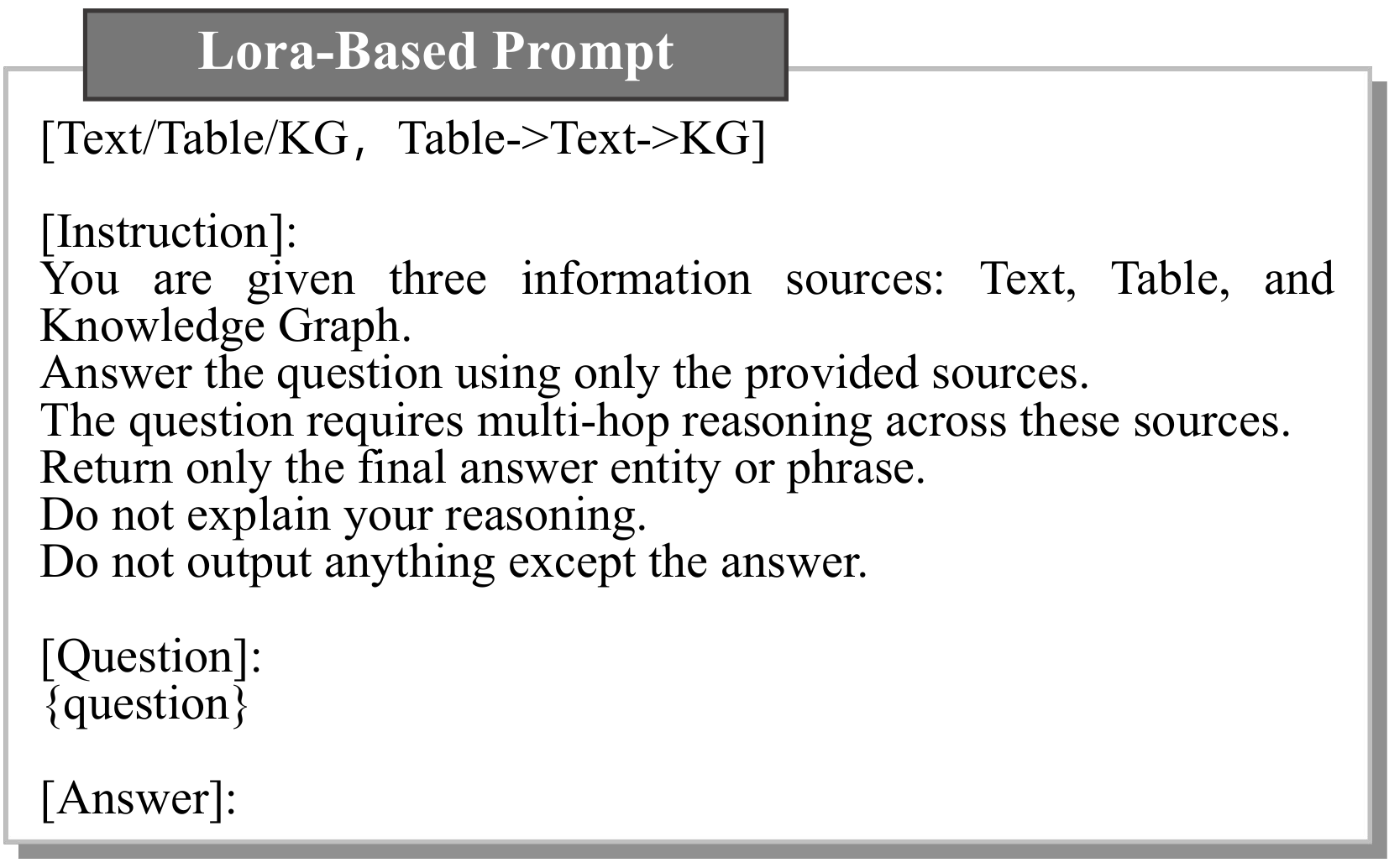}

        \captionof{figure}{
        Standard inference prompt.
        }
        \label{fig:lora_based_prompt}
    \end{minipage}
    \hfill
    \begin{minipage}[t]{0.495\textwidth}
        \centering
        \includegraphics[
            width=\linewidth
        ]{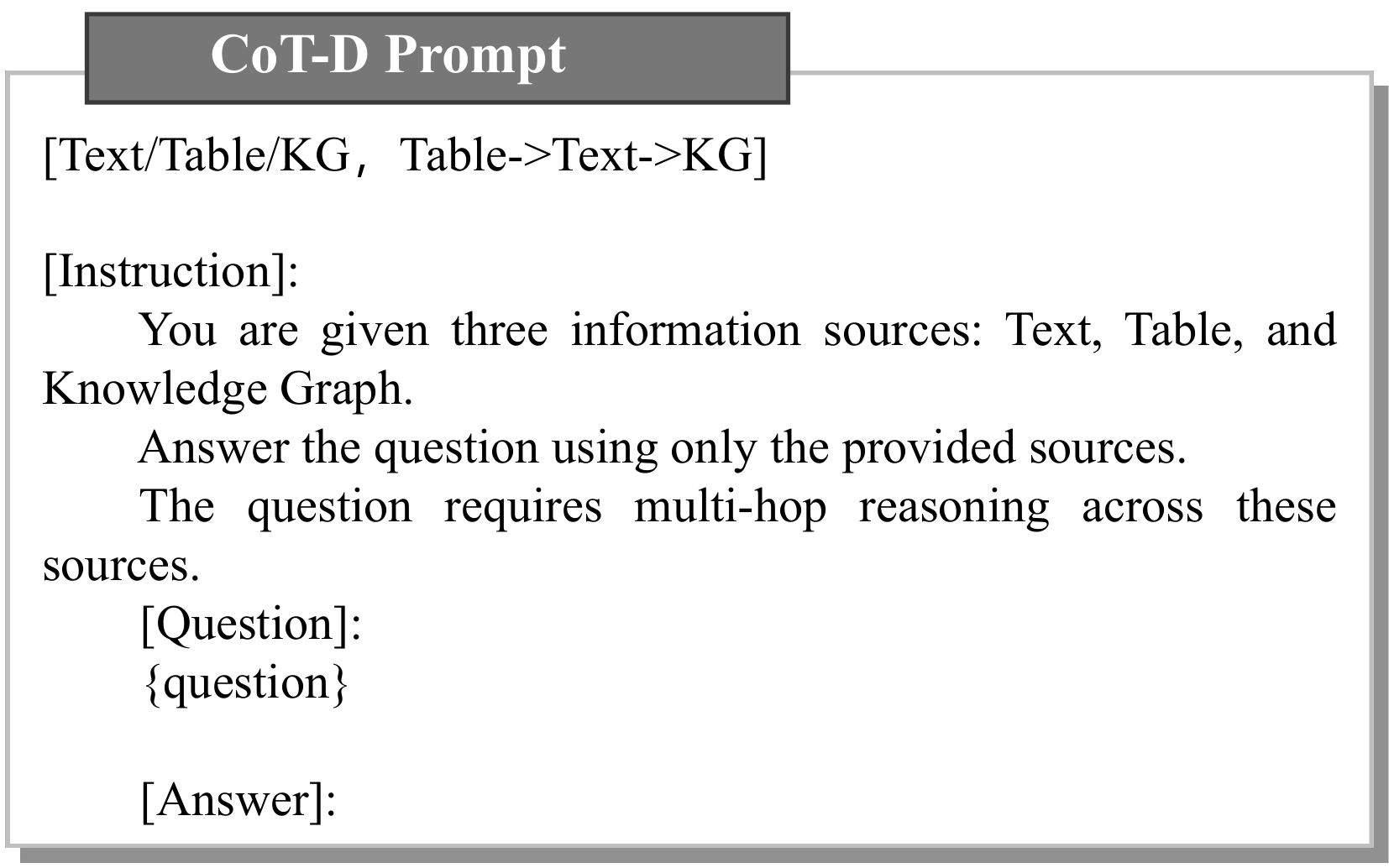}

        \captionof{figure}{
        CoT-D inference prompt.
        }
        \label{fig:cot_decoding_prompt}
    \end{minipage}

\end{figure*}

Figure~\ref{fig:cot_decoding_prompt} presents the prompt used for the
CoT-D baseline. 

\subsection{ReAct Prompt}
\label{app:react_prompt}

Figure~\ref{fig:ReAct_prompt} presents the prompt used for the ReAct
baseline. The model is instructed to iteratively reason over the table,
textual passage, and KG subgraph before producing the final answer. Its
intermediate reasoning is organized according to the prescribed
ReAct-style format.

\subsection{RCA Prompt}
\label{app:rca}

\begin{figure*}[!t]
    \centering

    \begin{minipage}[t]{0.495\textwidth}
        \centering
        \includegraphics[
            width=\linewidth
        ]{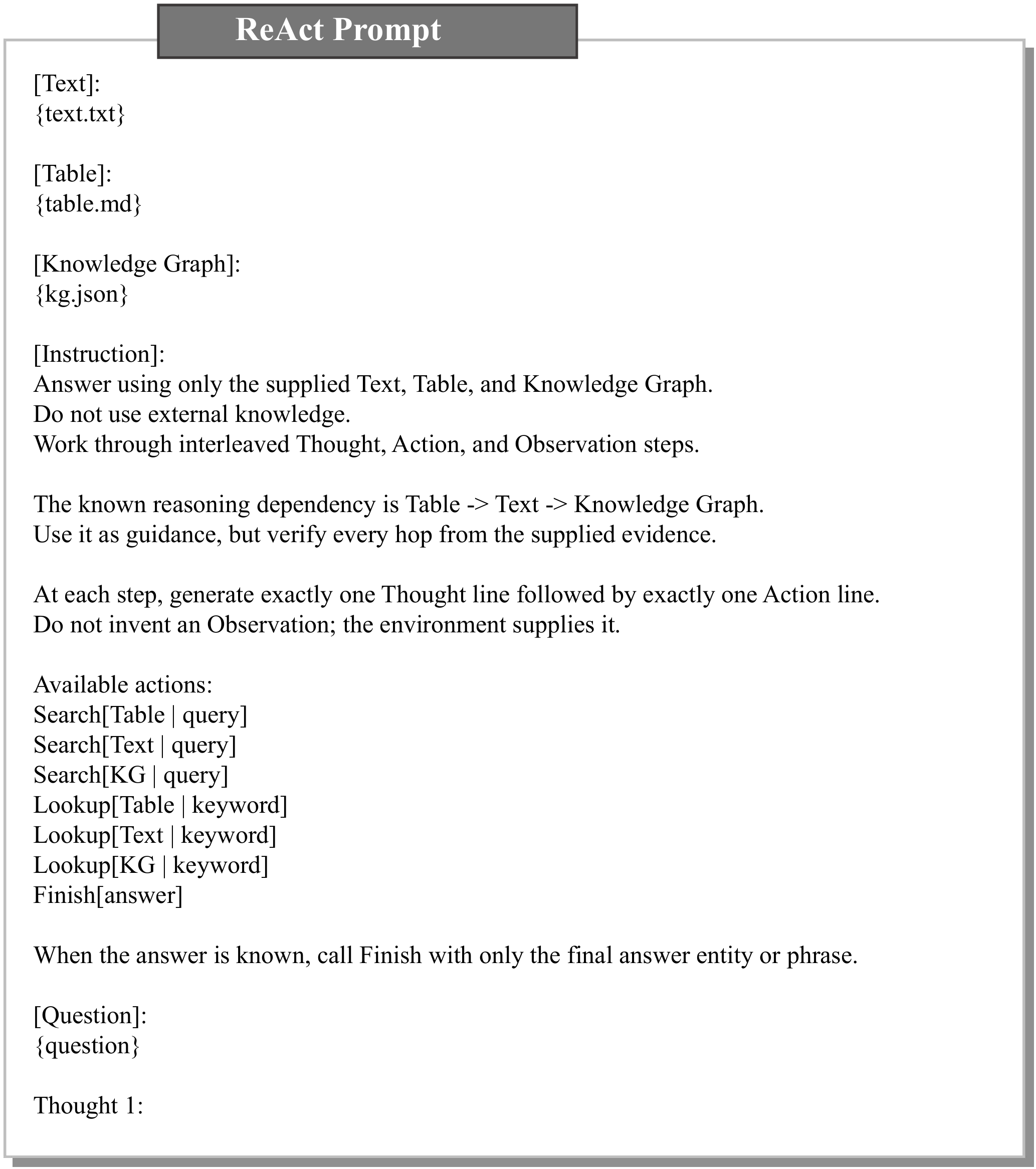}

        \captionof{figure}{
        ReAct inference prompt.
        }
        \label{fig:ReAct_prompt}
    \end{minipage}
    \hfill
    \begin{minipage}[t]{0.495\textwidth}
        \centering
        \includegraphics[
            width=\linewidth
        ]{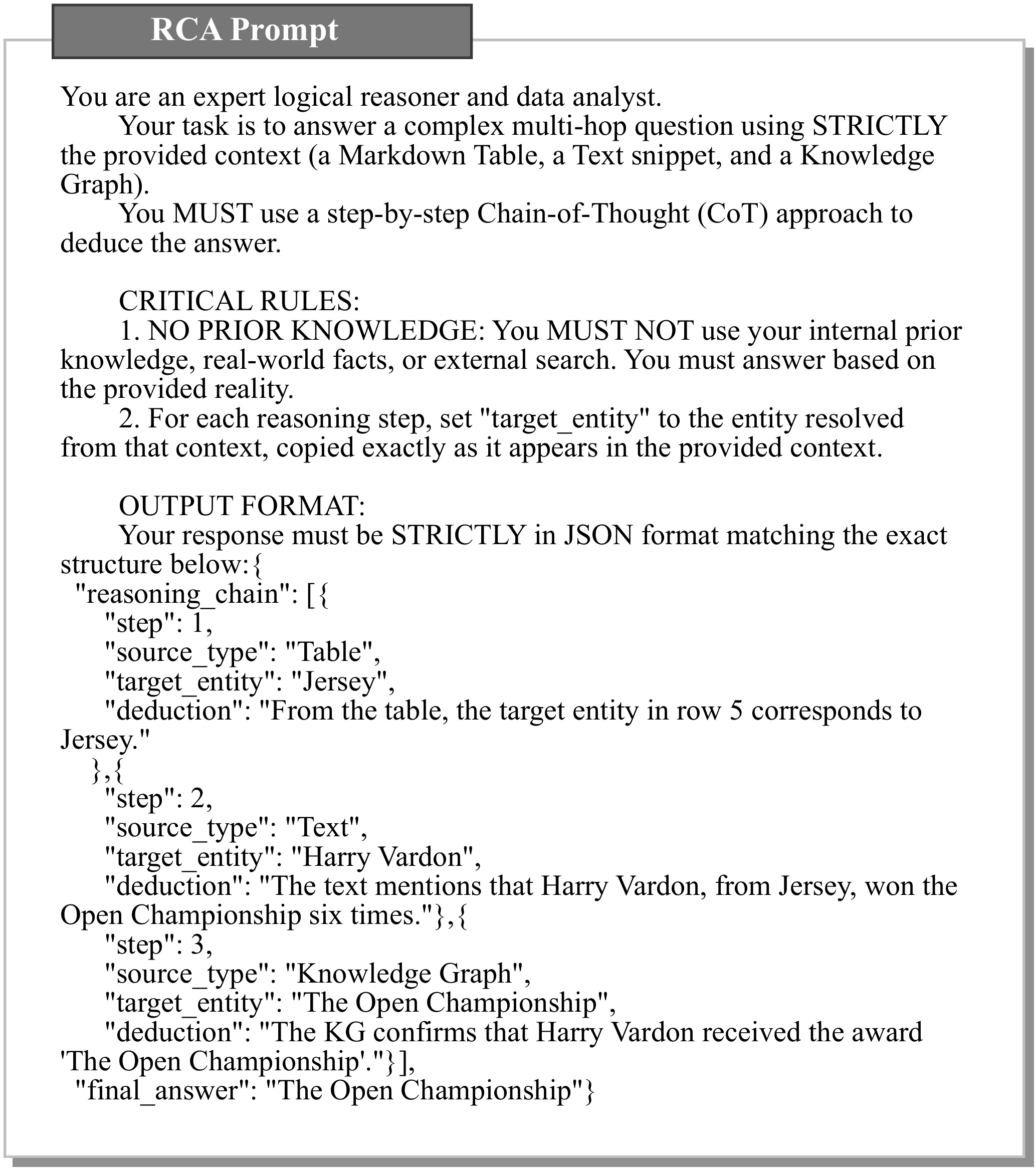}

        \captionof{figure}{
        Reasoning-chain elicitation prompt for RCA evaluation.
        }
        \label{fig:RCA_prompt}
    \end{minipage}

\end{figure*}

Figure~\ref{fig:RCA_prompt} presents the prompt used to elicit an
explicit reasoning chain for RCA evaluation. Given the serialized
heterogeneous knowledge contexts and the question, the model is
instructed to identify the relevant intermediate entities, describe
the reasoning process step by step, and provide the final answer in a
predefined format. The generated reasoning chain is then compared with
the annotated counterfactual reasoning chain to compute RCA.

%% file: custom.bib
@inproceedings{niu2024ragtruth,
  author    = {Cheng Niu and Yuanhao Wu and Juno Zhu and Siliang Xu and KaShun Shum and Randy Zhong and Juntong Song and Tong Zhang},
  title     = {{RAGTruth}: A Hallucination Corpus for Developing Trustworthy Retrieval-Augmented Language Models},
  booktitle = {Proceedings of the 62nd Annual Meeting of the Association for Computational Linguistics (Volume 1: Long Papers)},
  pages     = {10862--10878},
  year      = {2024},
  address   = {Bangkok, Thailand},
  publisher = {Association for Computational Linguistics},
  doi       = {10.18653/v1/2024.acl-long.585},
  url       = {https://aclanthology.org/2024.acl-long.585/}
}

@article{jacovi2025facts,
  author  = {Alon Jacovi and Andrew Wang and Chris Alberti and Connie Tao and Jon Lipovetz and Kate Olszewska and Lukas Haas and Michelle Liu and Nate Keating and Adam Bloniarz and Carl Saroufim and Corey Fry and Dror Marcus and Doron Kukliansky and Gaurav Singh Tomar and James Swirhun and Jinwei Xing and Lily Wang and Madhu Gurumurthy and Michael Aaron and Moran Ambar and Rachana Fellinger and Rui Wang and Zizhao Zhang and Sasha Goldshtein and Dipanjan Das},
  title   = {The {FACTS} Grounding Leaderboard: Benchmarking {LLM}s' Ability to Ground Responses to Long-Form Input},
  journal = {arXiv preprint arXiv:2501.03200},
  year    = {2025},
  doi     = {10.48550/arXiv.2501.03200},
  url     = {https://arxiv.org/abs/2501.03200}
}

@inproceedings{li2023halueval,
  author    = {Junyi Li and Xiaoxue Cheng and Xin Zhao and Jian-Yun Nie and Ji-Rong Wen},
  title     = {{HaluEval}: A Large-Scale Hallucination Evaluation Benchmark for Large Language Models},
  booktitle = {Proceedings of the 2023 Conference on Empirical Methods in Natural Language Processing},
  pages     = {6449--6464},
  year      = {2023},
  address   = {Singapore},
  publisher = {Association for Computational Linguistics},
  doi       = {10.18653/v1/2023.emnlp-main.397},
  url       = {https://aclanthology.org/2023.emnlp-main.397/}
}

@inproceedings{min2023factscore,
  author    = {Sewon Min and Kalpesh Krishna and Xinxi Lyu and Mike Lewis and Wen-tau Yih and Pang Wei Koh and Mohit Iyyer and Luke Zettlemoyer and Hannaneh Hajishirzi},
  title     = {{FActScore}: Fine-Grained Atomic Evaluation of Factual Precision in Long Form Text Generation},
  booktitle = {Proceedings of the 2023 Conference on Empirical Methods in Natural Language Processing},
  pages     = {12076--12100},
  year      = {2023},
  address   = {Singapore},
  publisher = {Association for Computational Linguistics},
  doi       = {10.18653/v1/2023.emnlp-main.741},
  url       = {https://aclanthology.org/2023.emnlp-main.741/}
}

@inproceedings{wang2024chain,
  author    = {Xuezhi Wang and Denny Zhou},
  title     = {Chain-of-Thought Reasoning without Prompting},
  booktitle = {Advances in Neural Information Processing Systems},
  volume    = {37},
  pages     = {66383--66409},
  year      = {2024},
  doi       = {10.52202/079017-2123},
  url       = {https://proceedings.neurips.cc/paper_files/paper/2024/hash/7a8e7fd295aa04eac4b470ae27f8785c-Abstract-Conference.html}
}

@inproceedings{yao2022react,
  author    = {Shunyu Yao and Jeffrey Zhao and Dian Yu and Nan Du and Izhak Shafran and Karthik R. Narasimhan and Yuan Cao},
  title     = {{ReAct}: Synergizing Reasoning and Acting in Language Models},
  booktitle = {The Eleventh International Conference on Learning Representations},
  year      = {2023},
  url       = {https://openreview.net/forum?id=WE_vluYUL-X}
}

@inproceedings{zhao2020transformer,
  author    = {Chen Zhao and Chenyan Xiong and Corby Rosset and Xia Song and Paul Bennett and Saurabh Tiwary},
  title     = {{Transformer-XH}: Multi-Evidence Reasoning with Extra Hop Attention},
  booktitle = {International Conference on Learning Representations},
  year      = {2020},
  url       = {https://openreview.net/forum?id=r1eIiCNYwS}
}

@inproceedings{huang2025masking,
  author    = {Wenyu Huang and Pavlos Vougiouklis and Mirella Lapata and Jeff Z. Pan},
  title     = {Masking in Multi-Hop {QA}: An Analysis of How Language Models Perform with Context Permutation},
  booktitle = {Proceedings of the 63rd Annual Meeting of the Association for Computational Linguistics (Volume 1: Long Papers)},
  pages     = {17781--17795},
  year      = {2025},
  address   = {Vienna, Austria},
  publisher = {Association for Computational Linguistics},
  doi       = {10.18653/v1/2025.acl-long.869},
  url       = {https://aclanthology.org/2025.acl-long.869/}
}

@inproceedings{wu2025cofca,
  author    = {Jian Wu and Linyi Yang and Zhen Wang and Manabu Okumura and Yue Zhang},
  title     = {{CofCA}: A Step-Wise Counterfactual Multi-Hop {QA} Benchmark},
  booktitle = {International Conference on Learning Representations},
  year      = {2025},
  url       = {https://proceedings.iclr.cc/paper_files/paper/2025/hash/2628d4d3b054c2d7ad33ab03435204f4-Abstract-Conference.html}
}

@inproceedings{chen2020hybridqa,
  author    = {Wenhu Chen and Hanwen Zha and Zhiyu Chen and Wenhan Xiong and Hong Wang and William Yang Wang},
  title     = {{HybridQA}: A Dataset of Multi-Hop Question Answering over Tabular and Textual Data},
  booktitle = {Findings of the Association for Computational Linguistics: EMNLP 2020},
  pages     = {1026--1036},
  year      = {2020},
  address   = {Online},
  publisher = {Association for Computational Linguistics},
  doi       = {10.18653/v1/2020.findings-emnlp.91},
  url       = {https://aclanthology.org/2020.findings-emnlp.91/}
}

@article{egressy2025setllm,
  author  = {Beni Egressy and Jan St{\"u}hmer},
  title   = {{Set-LLM}: A Permutation-Invariant {LLM}},
  journal = {arXiv preprint arXiv:2505.15433},
  year    = {2025},
  doi     = {10.48550/arXiv.2505.15433},
  url     = {https://arxiv.org/abs/2505.15433}
}

@article{su2021roformer,
  author  = {Jianlin Su and Murtadha H. M. Ahmed and Yu Lu and Shengfeng Pan and Bo Wen and Yunfeng Liu},
  title   = {{RoFormer}: Enhanced Transformer with Rotary Position Embedding},
  journal = {Neurocomputing},
  volume  = {568},
  pages   = {127063},
  year    = {2024},
  doi     = {10.1016/j.neucom.2023.127063},
  url     = {https://doi.org/10.1016/j.neucom.2023.127063}
}

@inproceedings{haviv2022transformer,
  author    = {Adi Haviv and Ori Ram and Ofir Press and Peter Izsak and Omer Levy},
  title     = {Transformer Language Models without Positional Encodings Still Learn Positional Information},
  booktitle = {Findings of the Association for Computational Linguistics: EMNLP 2022},
  pages     = {1382--1390},
  year      = {2022},
  address   = {Abu Dhabi, United Arab Emirates},
  publisher = {Association for Computational Linguistics},
  doi       = {10.18653/v1/2022.findings-emnlp.99},
  url       = {https://aclanthology.org/2022.findings-emnlp.99/}
}

@article{vajda2026teaching,
  author  = {Dario Vajda},
  title   = {Teaching {LLM}s to See Graphs: Unifying Text and Structural Reasoning},
  journal = {arXiv preprint arXiv:2605.10247},
  year    = {2026},
  doi     = {10.48550/arXiv.2605.10247},
  url     = {https://arxiv.org/abs/2605.10247}
}

@inproceedings{zhang2021magnet,
  author    = {Xitong Zhang and Yixuan He and Nathan Brugnone and Michael Perlmutter and Matthew Hirn},
  title     = {{MagNet}: A Neural Network for Directed Graphs},
  booktitle = {Advances in Neural Information Processing Systems},
  volume    = {34},
  pages     = {27003--27015},
  year      = {2021},
  url       = {https://proceedings.neurips.cc/paper_files/paper/2021/hash/e32084632d369461572832e6582aac36-Abstract.html}
}

@inproceedings{yang2018hotpotqa,
  author    = {Zhilin Yang and Peng Qi and Saizheng Zhang and Yoshua Bengio and William Cohen and Ruslan Salakhutdinov and Christopher D. Manning},
  title     = {{HotpotQA}: A Dataset for Diverse, Explainable Multi-Hop Question Answering},
  booktitle = {Proceedings of the 2018 Conference on Empirical Methods in Natural Language Processing},
  pages     = {2369--2380},
  year      = {2018},
  address   = {Brussels, Belgium},
  publisher = {Association for Computational Linguistics},
  doi       = {10.18653/v1/D18-1259},
  url       = {https://aclanthology.org/D18-1259/}
}

@article{vrandecic2014wikidata,
  author    = {Denny Vrande{\v{c}}i{\'{c}} and Markus Kr{\"o}tzsch},
  title     = {{Wikidata}: A Free Collaborative Knowledgebase},
  journal   = {Communications of the ACM},
  volume    = {57},
  number    = {10},
  pages     = {78--85},
  year      = {2014},
  month     = oct,
  publisher = {Association for Computing Machinery},
  doi       = {10.1145/2629489},
  url       = {https://doi.org/10.1145/2629489}
}

@inproceedings{tang2024minicheck,
  author    = {Liyan Tang and Philippe Laban and Greg Durrett},
  title     = {{MiniCheck}: Efficient Fact-Checking of {LLM}s on Grounding Documents},
  booktitle = {Proceedings of the 2024 Conference on Empirical Methods in Natural Language Processing},
  month     = nov,
  year      = {2024},
  address   = {Miami, Florida, USA},
  publisher = {Association for Computational Linguistics},
  pages     = {8818--8847},
  doi       = {10.18653/v1/2024.emnlp-main.499},
  url       = {https://aclanthology.org/2024.emnlp-main.499/}
}

@inproceedings{xu2024pride,
  author    = {Wenda Xu and Guanglei Zhu and Xuandong Zhao and Liangming Pan and Lei Li and William Yang Wang},
  title     = {Pride and Prejudice: {LLM} Amplifies Self-Bias in Self-Refinement},
  booktitle = {Proceedings of the 62nd Annual Meeting of the Association for Computational Linguistics (Volume 1: Long Papers)},
  month     = aug,
  year      = {2024},
  address   = {Bangkok, Thailand},
  publisher = {Association for Computational Linguistics},
  pages     = {15474--15492},
  doi       = {10.18653/v1/2024.acl-long.826},
  url       = {https://aclanthology.org/2024.acl-long.826/}
}

@inproceedings{liu2025grl,
  author    = {Yuze Liu and Tingjie Liu and Tiehua Zhang and Youhua Xia and Jinze Wang and Zhishu Shen and Jiong Jin and Zhijun Ding and Fei Richard Yu},
  title     = {{GRL-Prompt}: Towards Prompts Optimization via Graph-Empowered Reinforcement Learning Using {LLM}s' Feedback},
  booktitle = {Data Science: Foundations and Applications -- 29th Pacific-Asia Conference on Knowledge Discovery and Data Mining, PAKDD 2025, Proceedings, Part VII},
  series    = {Lecture Notes in Computer Science},
  volume    = {15876},
  pages     = {426--438},
  year      = {2025},
  publisher = {Springer},
  doi       = {10.1007/978-981-96-8298-0_34},
  url       = {https://doi.org/10.1007/978-981-96-8298-0_34}
}

@inproceedings{liu2026structure,
  author    = {Yuze Liu and Yunhan Wang and Tiehua Zhang and Zhishu Shen and Cheng Peng and Libing Wu and Feng Xia and Jiong Jin},
  title     = {A Structure-Agnostic Co-Tuning Framework for {LLM}s and {SLM}s in Cloud-Edge Systems},
  booktitle = {Proceedings of the ACM Web Conference 2026},
  pages     = {5667--5675},
  year      = {2026},
  publisher = {Association for Computing Machinery},
  doi       = {10.1145/3774904.3792682},
  url       = {https://doi.org/10.1145/3774904.3792682}
}

@inproceedings{jiangdivide,
  author    = {Yigeng Jiang and Tingjun Su and Tong Wu and Shumeng Zhang and Yanxu Zhao and Zhaohong Huang and Tianyu Xie and Yuhang Wu and Yisheng Lin and Yuze Liu and others},
  title     = {Divide, Verify, and Conquer: Verification-Guided {DAG} Reasoning for Large Language Models},
  booktitle = {2026 International Joint Conference on Neural Networks},
  year      = {2026},
  url       = {https://linklings.s3.amazonaws.com/organizations/WCCI/wcci2026/submissions/stype114/Ycvt1-ijcnn_pap4933s2.pdf},
  note      = {Paper 4933}
}

@article{yang2025qwen3,
  title   = {{Qwen3} Technical Report},
  author  = {Yang, An and Li, Anfeng and Yang, Baosong and others},
  journal = {arXiv preprint arXiv:2505.09388},
  year    = {2025},
  doi     = {10.48550/arXiv.2505.09388},
  url     = {https://arxiv.org/abs/2505.09388}
}

@article{gemmateam2025gemma3,
  title   = {{Gemma 3} Technical Report},
  author  = {{Gemma Team}},
  journal = {arXiv preprint arXiv:2503.19786},
  year    = {2025},
  doi     = {10.48550/arXiv.2503.19786},
  url     = {https://arxiv.org/abs/2503.19786}
}

@article{minimax2026m2,
  title   = {The {MiniMax-M2} Series: Mini Activations Unleashing
             Max Real-World Intelligence},
  author  = {{MiniMax}},
  journal = {arXiv preprint arXiv:2605.26494},
  year    = {2026},
  doi     = {10.48550/arXiv.2605.26494},
  url     = {https://arxiv.org/abs/2605.26494}
}

@article{kimiteam2026kimik25,
  title   = {{Kimi K2.5}: Visual Agentic Intelligence},
  author  = {{Kimi Team}},
  journal = {arXiv preprint arXiv:2602.02276},
  year    = {2026},
  doi     = {10.48550/arXiv.2602.02276},
  url     = {https://arxiv.org/abs/2602.02276}
}

@article{deepseekai2025v32,
  title   = {{DeepSeek-V3.2}: Pushing the Frontier of Open Large
             Language Models},
  author  = {{DeepSeek-AI}},
  journal = {arXiv preprint arXiv:2512.02556},
  year    = {2025},
  doi     = {10.48550/arXiv.2512.02556},
  url     = {https://arxiv.org/abs/2512.02556}
}

@misc{deepseekai2026v4,
  author = {{DeepSeek-AI}},
  title  = {{DeepSeek-V4}: Towards Highly Efficient Million-Token Context Intelligence},
  year   = {2026},
  url    = {https://huggingface.co/deepseek-ai/DeepSeek-V4-Flash},
  note   = {Technical report}
}

@article{geminiteam2025gemini25,
  title   = {{Gemini 2.5}: Pushing the Frontier with Advanced
             Reasoning, Multimodality, Long Context, and Next
             Generation Agentic Capabilities},
  author  = {{Gemini Team}},
  journal = {arXiv preprint arXiv:2507.06261},
  year    = {2025},
  doi     = {10.48550/arXiv.2507.06261},
  url     = {https://arxiv.org/abs/2507.06261}
}

@article{teamglm2024chatglm,
  title   = {{ChatGLM}: A Family of Large Language Models from
             {GLM-130B} to {GLM-4} All Tools},
  author  = {{Team GLM}},
  journal = {arXiv preprint arXiv:2406.12793},
  year    = {2024},
  doi     = {10.48550/arXiv.2406.12793},
  url     = {https://arxiv.org/abs/2406.12793}
}

@article{grattafiori2024llama3,
  title   = {The {Llama 3} Herd of Models},
  author  = {Grattafiori, Aaron and others},
  journal = {arXiv preprint arXiv:2407.21783},
  year    = {2024},
  doi     = {10.48550/arXiv.2407.21783},
  url     = {https://arxiv.org/abs/2407.21783}
}

@article{jiang2023mistral,
  title   = {{Mistral 7B}},
  author  = {Jiang, Albert Q. and Sablayrolles, Alexandre and
             Mensch, Arthur and others},
  journal = {arXiv preprint arXiv:2310.06825},
  year    = {2023},
  doi     = {10.48550/arXiv.2310.06825},
  url     = {https://arxiv.org/abs/2310.06825}
}

@inproceedings{chia-etal-2025-longdoc,
    title = "{M}-{L}ong{D}oc: A Benchmark For Multimodal Super-Long Document Understanding And A Retrieval-Aware Tuning Framework",
    author = "Chia, Yew Ken  and
      Cheng, Liying  and
      Chan, Hou Pong  and
      Song, Maojia  and
      Liu, Chaoqun  and
      Aljunied, Mahani  and
      Poria, Soujanya  and
      Bing, Lidong",
    editor = "Christodoulopoulos, Christos  and
      Chakraborty, Tanmoy  and
      Rose, Carolyn  and
      Peng, Violet",
    booktitle = "Proceedings of the 2025 Conference on Empirical Methods in Natural Language Processing",
    month = nov,
    year = "2025",
    address = "Suzhou, China",
    publisher = "Association for Computational Linguistics",
    url = "https://aclanthology.org/2025.emnlp-main.469/",
    doi = "10.18653/v1/2025.emnlp-main.469",
    pages = "9233--9250",
    isbn = "979-8-89176-332-6"
}

@article{al2026dagger,
  author  = {Zabir Al Nazi and Shubhashis Roy Dipta and Sudipta Kar},
  title   = {{\textdagger DAGGER}: Distractor-Aware Graph Generation for Executable Reasoning in Math Problems},
  journal = {arXiv preprint arXiv:2601.06853},
  year    = {2026},
  doi     = {10.48550/arXiv.2601.06853},
  url     = {https://arxiv.org/abs/2601.06853}
}

@inproceedings{liang2022astbert,
  author    = {Rong Liang and Tiehua Zhang and Yujie Lu and Yuze Liu and Zhen Huang and Xin Chen},
  title     = {{AstBERT}: Enabling Language Model for Financial Code Understanding with Abstract Syntax Trees},
  booktitle = {Proceedings of the Fourth Workshop on Financial Technology and Natural Language Processing (FinNLP)},
  month     = dec,
  year      = {2022},
  address   = {Abu Dhabi, United Arab Emirates (Hybrid)},
  publisher = {Association for Computational Linguistics},
  pages     = {10--17},
  doi       = {10.18653/v1/2022.finnlp-1.2},
  url       = {https://aclanthology.org/2022.finnlp-1.2/}
}

@inproceedings{chen2026tasks,
  author    = {Zichen Chen and Jianda Chen and Jiaao Chen and Misha Sra},
  title     = {From Tasks to Teams: A Risk-First Evaluation Framework for Multi-Agent {LLM} Systems in Finance},
  booktitle = {Findings of the Association for Computational Linguistics: ACL 2026},
  month     = jul,
  year      = {2026},
  address   = {San Diego, California, United States},
  publisher = {Association for Computational Linguistics},
  pages     = {38819--38857},
  doi       = {10.18653/v1/2026.findings-acl.1934},
  url       = {https://aclanthology.org/2026.findings-acl.1934/}
}

@inproceedings{zhang2026medtvt,
  author    = {Yuting Zhang and Kaishen Yuan and Hao Lu and Yutao Yue and Jintai Chen and Kaishun Wu},
  title     = {{MedTVT-R1}: A Multimodal {LLM} Empowering Medical Reasoning and Diagnosis},
  booktitle = {Proceedings of the IEEE/CVF Conference on Computer Vision and Pattern Recognition},
  month     = jun,
  year      = {2026},
  pages     = {35248--35259}
}

@article{liu2025comprehensive,
  author  = {Jiaheng Liu and Dawei Zhu and Zhiqi Bai and Yancheng He and Huanxuan Liao and Haoran Que and Zekun Wang and Chenchen Zhang and Ge Zhang and Jiebin Zhang and others},
  title   = {A Comprehensive Survey on Long-Context Language Modeling},
  journal = {arXiv preprint arXiv:2503.17407},
  year    = {2025},
  doi     = {10.48550/arXiv.2503.17407},
  url     = {https://arxiv.org/abs/2503.17407}
}

@article{zhao2026retrieval,
  author    = {Penghao Zhao and Hailin Zhang and Qinhan Yu and Zhengren Wang and Yunteng Geng and Fangcheng Fu and Ling Yang and Wentao Zhang and Jie Jiang and Bin Cui},
  title     = {Retrieval-Augmented Generation for {AI}-Generated Content: A Survey},
  journal   = {Data Science and Engineering},
  volume    = {11},
  number    = {1},
  pages     = {1--29},
  year      = {2026},
  publisher = {Springer},
  doi       = {10.1007/s41019-025-00335-5},
  url       = {https://doi.org/10.1007/s41019-025-00335-5}
}

@inproceedings{lei2023s3hqa,
  author    = {Fangyu Lei and Xiang Li and Yifan Wei and Shizhu He and Yiming Huang and Jun Zhao and Kang Liu},
  title     = {{S3HQA}: A Three-Stage Approach for Multi-Hop Text-Table Hybrid Question Answering},
  booktitle = {Proceedings of the 61st Annual Meeting of the Association for Computational Linguistics (Volume 2: Short Papers)},
  month     = jul,
  year      = {2023},
  address   = {Toronto, Canada},
  publisher = {Association for Computational Linguistics},
  pages     = {1731--1740},
  doi       = {10.18653/v1/2023.acl-short.147},
  url       = {https://aclanthology.org/2023.acl-short.147/}
}

@inproceedings{afzal2025knowing,
  author    = {Anum Afzal and Florian Matthes and Gal Chechik and Yftah Ziser},
  title     = {Knowing Before Saying: {LLM} Representations Encode Information About Chain-of-Thought Success Before Completion},
  booktitle = {Findings of the Association for Computational Linguistics: ACL 2025},
  month     = jul,
  year      = {2025},
  address   = {Vienna, Austria},
  publisher = {Association for Computational Linguistics},
  pages     = {12791--12806},
  doi       = {10.18653/v1/2025.findings-acl.662},
  url       = {https://aclanthology.org/2025.findings-acl.662/}
}

@article{guan2025order,
  author  = {Bryan Guan and Tanya Roosta and Peyman Passban and Mehdi Rezagholizadeh},
  title   = {The Order Effect: Investigating Prompt Sensitivity in Closed-Source {LLM}s},
  journal = {arXiv preprint arXiv:2502.04134},
  year    = {2025},
  doi     = {10.48550/arXiv.2502.04134},
  url     = {https://arxiv.org/abs/2502.04134}
}

@article{liu2026ml,
  title={ML-ECS: A Collaborative Multimodal Learning Framework for Edge-Cloud Synergies},
  author={Liu, Yuze and Chu, Shibo and Zhang, Tiehua and Zhou, Hao and Shen, Zhishu and Wang, Jinze and Qi, Jianzhong and Xia, Feng},
  journal={arXiv preprint arXiv:2602.14107},
  year={2026}
}
